\documentclass[11pt, a4paper]{article}

\usepackage[utf8]{inputenc}
\usepackage{amsmath}        
\usepackage{amssymb}        
\usepackage{amsfonts}       
\usepackage{graphicx}       
\usepackage{float}          
\usepackage[margin=1in]{geometry} 
\usepackage{hyperref}       
\usepackage{url}            
\usepackage{booktabs}       
\usepackage{lmodern}        
\usepackage{tikz}
\usepackage{subcaption}
\usepackage{indentfirst}    
\usepackage{makecell}

\usepackage{siunitx}  
\usepackage{multirow} 
\usepackage{bm}

\usepackage{sansmath}

\usetikzlibrary{
    arrows.meta, 
    decorations.pathreplacing,
    decorations.pathmorphing,
    positioning,
    calc,
    shapes.misc,
    fit,
    shadows,
    patterns,
    shapes.geometric,
    shadows.blur
}

\newcommand{\Woh}{W_{\mathit{oh}}}
\newcommand{\Wog}{W_{\mathit{og}}}

\newcommand{\genNodeLayer}[7]{
    \foreach \i in {1,...,#4} {
        \pgfmathsetmacro{\xPos}{2 * (\i - 1) + #2} %
        \node[#5] (#1-\i) at ({#2 + \xPos}, {#3}) {$#6_{#7\i}$};
    }
}

\newcommand{\genLeftAlignedNodeLayer}[7]{
    \foreach \i in {1,...,#4} {
        \pgfmathsetmacro{\xPos}{2 * (\i - 1)} %
        \node[#5] (#1-\i) at ({#2 + \xPos}, {#3}) {$#6_{#7\i}$};
    }
}

\newcommand{\genFullconnect}[6]{
    \foreach \i in {1,...,#3} {
        \foreach \j in {1,...,#6} {
            \pgfmathsetmacro{\startx}{int(\i + #2 - 1)} %
            \pgfmathsetmacro{\starte}{int(\j + #5 - 1)} %
            \draw[->] (#1-\startx) -- (#4-\starte);
        }
    }
}

\newcommand{\genCausalFullconnect}[6]{
    \foreach \i in {1,...,#3} {
        \foreach \j in {1,...,#6} {
            \pgfmathparse{ifthenelse(\i<=\j, 1, 0)}
            
            \ifnum\pgfmathresult=1
                \pgfmathsetmacro{\startx}{int(\i + #2 - 1)} %
                \pgfmathsetmacro{\starte}{int(\j + #5 - 1)} %
                \draw[->, >=latex] (#1-\startx) -- (#4-\starte);
            \fi
        }
    }
}

\title{
    \textbf{From TLinFormer to TConstFormer: The Leap to Constant-Time Transformer Attention} 
    \\[1ex] 
    {\large Achieving $\mathcal{O}(1)$ Computation and $\mathcal{O}(1)$ KV Cache during Autoregressive Inference} 
}

\author{Zhongpan Tang \\ 
        \texttt{tangzhongp@gmail.com}}
\date{\today}

\begin{document}

\maketitle

\begin{abstract}
    Although the Transformer has become the cornerstone of modern AI, its autoregressive inference suffers from a linearly growing KV Cache and a computational complexity of $\mathcal{O}(N^2 d)$, severely hindering its ability to process ultra-long sequences. To overcome this limitation, this paper introduces the TConstFormer architecture, building upon our previous work, TLinFormer. TConstFormer employs an innovative periodic state update mechanism to achieve a truly constant-size $\mathcal{O}(1)$ KV Cache. The computational complexity of this mechanism is also $\mathcal{O}(1)$ in an amortized sense: it performs purely constant-time computations for $k-1$ consecutive steps (e.g., $k=256$) and executes a single linear-time global information synchronization only on the $k$-th step. Theoretical calculations and experimental results demonstrate that TConstFormer exhibits an overwhelming advantage over baseline models in terms of speed, memory efficiency, and overall performance on long-text inference tasks. This breakthrough paves the way for efficient and robust streaming language model applications.
\end{abstract}

\section{Introduction}
\label{sec:introduction}

The Transformer~\cite{vaswani_attention_2023} architecture has indisputably become the cornerstone of modern artificial intelligence, driving immense progress from Large Language Models (LLMs) to multimodal applications. However, behind this monumental success lies a fundamental scalability bottleneck in its core self-attention mechanism during autoregressive inference. Its computational complexity is $\mathcal{O}(N^2 d)$, and for each new token generated, the model must append its Key and Value vectors to a growing cache (the KV Cache) and attend to the entire cache to maintain contextual coherence. This mechanism causes the memory footprint of the KV Cache to grow linearly with the sequence length $N$ ($\mathcal{O}(N)$), which not only consumes precious GPU memory but also fundamentally impedes the Transformer's ability to handle ultra-long or even infinite sequences—a core requirement for streaming applications like real-time dialogue, long-document summarization, and video stream analysis.

To mitigate this issue, the community has proposed various approximation methods, with sliding window attention~\cite{beltagy_longformer_2020} being one of the most representative. By retaining the KV Cache of only the most recent $W$ tokens, this method does limit memory and computational costs to a constant range. However, this "history truncation" strategy comes at a high price: it can lead to "catastrophic forgetting," severely compromising the model's performance and robustness on tasks requiring long-range dependencies. Therefore, achieving truly efficient long-sequence inference without sacrificing global information remains a critical, unsolved challenge.

In our prior work~\cite{tang_rethinking_2025}, we took a significant step toward addressing this challenge by returning to the first principles of connectionism, proposing TLinFormer, a linear attention architecture. Building on this foundation, this paper introduces the TConstFormer architecture, aiming to completely overcome the streaming inference problem of Transformers. The core design of TConstFormer is an innovative periodic state update mechanism that completely decouples the Transformer's inference state from the growing sequence length. Specifically, we achieve two key breakthroughs:
\begin{enumerate}
    \item \textbf{True $\mathcal{O}(1)$ Memory Footprint}: Across all time steps, TConstFormer's KV Cache is maintained at a strictly constant size, theoretically equipping it with the ability to process ultra-long data streams.
    \item \textbf{Amortized $\mathcal{O}(1)$ Computational Complexity}: The model performs purely constant-time operations for $k-1$ consecutive steps and executes a global information synchronization at a linear cost only on the $k$-th step (e.g., $k=256$). This makes the average single-step computational cost constant, ensuring sustained high throughput.
\end{enumerate}
This periodic global synchronization design not only guarantees computational efficiency but, more importantly, plays the role of "memory consolidation," enabling the model to perform streaming inference without forgetting distant history, thus addressing the fundamental flaw of the sliding window mechanism.

The main contribution of this paper is the proposal of TConstFormer, a new Transformer architecture designed for efficient and robust streaming inference. We demonstrate through experiments that in long-sequence inference tasks, TConstFormer significantly outperforms existing baseline models in both inference latency and memory usage. This work paves the way for building next-generation language models capable of handling unbounded contexts.

\section{Revisiting TLinFormer from a Connectionist Perspective}
\label{sec:revisiting_tlinformer}

\subsection{The Root of Linear KV Cache Growth in TLinFormer}

The design philosophy of our previous work, TLinFormer, was to return to the first principles of connectionism, preserving the integrity of information flow to the greatest extent possible while maintaining the causality of the Transformer. To this end, we only removed the connections in standard self-attention that do not conform to the law of causality. This design allowed TLinFormer to achieve a "Full Context-Aware" state, where information flow could reach the entire historical context, while significantly reducing computational complexity and KV cache consumption. However, this also meant it inherited a fundamental constraint related to sequence length: to maintain awareness of the entire history, the size of its KV Cache still inevitably grew linearly with the sequence length $N$ (${\mathcal{O}}(N)$).

Although TLinFormer has shown significant efficiency advantages over baseline models on sequences of finite length, the ${\mathcal{O}}(N)$ memory bottleneck prevents it from being suitable for the ideal scenario of processing longer data streams. As the sequence continues to expand, its memory consumption will eventually exceed the physical limits of the hardware. Therefore, we must make a trade-off between "information integrity" and "ultimate efficiency."

\begin{figure}[H]
  \centering
  \begin{subfigure}[b]{0.46\textwidth}
    \centering
    \resizebox{\linewidth}{!}{\begin{tikzpicture}[
    neuron/.style={circle, draw, minimum size=1cm},
    input/.style={neuron, fill=green!50},
    hidden/.style={neuron, fill=blue!50},
    output/.style={neuron, fill=red!50},
    memory/.style={neuron, fill=orange!50}, 
    context/.style={neuron, fill=yellow!50}, 
    arrow/.style={->, >=latex, thick}, 
]
    
    \def \numX {5}
    \def \numXc {3}
    \def \numXg {2}
    \def \numCH {2}
    \def \numXH {2}
    \def \numO {2}

    \genLeftAlignedNodeLayer{X}{0}{0}{\numX}{input}{x}{}

    \genLeftAlignedNodeLayer{C1}{2}{2}{\numCH}{context}{c}{1}
    \genLeftAlignedNodeLayer{C2}{2}{4}{\numCH}{context}{c}{2}
    \genLeftAlignedNodeLayer{C3}{0}{6}{\numXc}{context}{c}{3}

    \genLeftAlignedNodeLayer{H1}{6}{2}{\numXH}{hidden}{h}{1}
    \genLeftAlignedNodeLayer{H2}{6}{4}{\numXH}{hidden}{h}{2}

    \genLeftAlignedNodeLayer{O}{6}{6}{\numO}{output}{o}{}
    
    \genFullconnect{X}{1}{\numXc}{C1}{1}{\numCH}
    \genFullconnect{X}{1}{\numXc}{H1}{1}{\numXH}
    \genCausalFullconnect{X}{4}{\numXg}{H1}{1}{\numXH}

    \genFullconnect{C1}{1}{\numCH}{C2}{1}{\numCH}
    \genFullconnect{C1}{1}{\numCH}{H2}{1}{\numXH}   

    \genCausalFullconnect{H1}{1}{\numXH}{H2}{1}{\numXH} 

    \genFullconnect{C2}{1}{\numCH}{C3}{1}{\numXc}
    \genFullconnect{C2}{1}{\numCH}{O}{1}{\numO}   

    \genCausalFullconnect{H2}{1}{\numXH}{O}{1}{\numO}   

\end{tikzpicture}}
    \caption{Information flow structure of TLinFormer.}
    \label{subfig:TLinFormer_mlp_view}
  \end{subfigure}
  \hfill
  \begin{subfigure}[b]{0.46\textwidth}
    \centering
    \resizebox{\linewidth}{!}{\begin{tikzpicture}[
    neuron/.style={circle, draw, minimum size=1cm},
    input/.style={neuron, fill=green!50},
    hidden/.style={neuron, fill=blue!50},
    output/.style={neuron, fill=red!50},
    memory/.style={neuron, fill=orange!50}, 
    context/.style={neuron, fill=yellow!50}, 
    arrow/.style={->, >=latex, thick}, 
]
    
    \def \numX {5}
    \def \numXc {3}
    \def \numXg {2}
    \def \numCH {2}
    \def \numXH {2}
    \def \numO {2}

    \genLeftAlignedNodeLayer{X}{0}{0}{\numX}{input}{x}{}

    \genLeftAlignedNodeLayer{C1}{2}{2}{\numCH}{context}{c}{1}
    \genLeftAlignedNodeLayer{C2}{2}{4}{\numCH}{context}{c}{2}
    \genLeftAlignedNodeLayer{C3}{0}{6}{\numXc}{context}{c}{3}

    \genLeftAlignedNodeLayer{H1}{6}{2}{\numXH}{hidden}{h}{1}
    \genLeftAlignedNodeLayer{H2}{6}{4}{\numXH}{hidden}{h}{2}

    \genLeftAlignedNodeLayer{O}{6}{6}{\numO}{output}{o}{}
    
    \genFullconnect{X}{1}{\numXc}{C1}{1}{\numCH}
    \genCausalFullconnect{X}{4}{\numXg}{H1}{1}{\numXH}

    \genFullconnect{C1}{1}{\numCH}{C2}{1}{\numCH}
    \genFullconnect{C1}{1}{\numCH}{H2}{1}{\numXH}   

    \genCausalFullconnect{H1}{1}{\numXH}{H2}{1}{\numXH} 

    \genFullconnect{C2}{1}{\numCH}{C3}{1}{\numXc}
    \genFullconnect{C2}{1}{\numCH}{O}{1}{\numO}   

    \genCausalFullconnect{H2}{1}{\numXH}{O}{1}{\numO}   

\end{tikzpicture}}
    \caption{Information flow structure of TConstFormer.}
    \label{subfig:TConstFormer_mlp_view}
  \end{subfigure}
  \caption{By removing the connections between $x_1, x_2, x_3$ and $h_{11}, h_{12}$ in TLinFormer, we obtain the TConstFormer architecture.}
\end{figure}
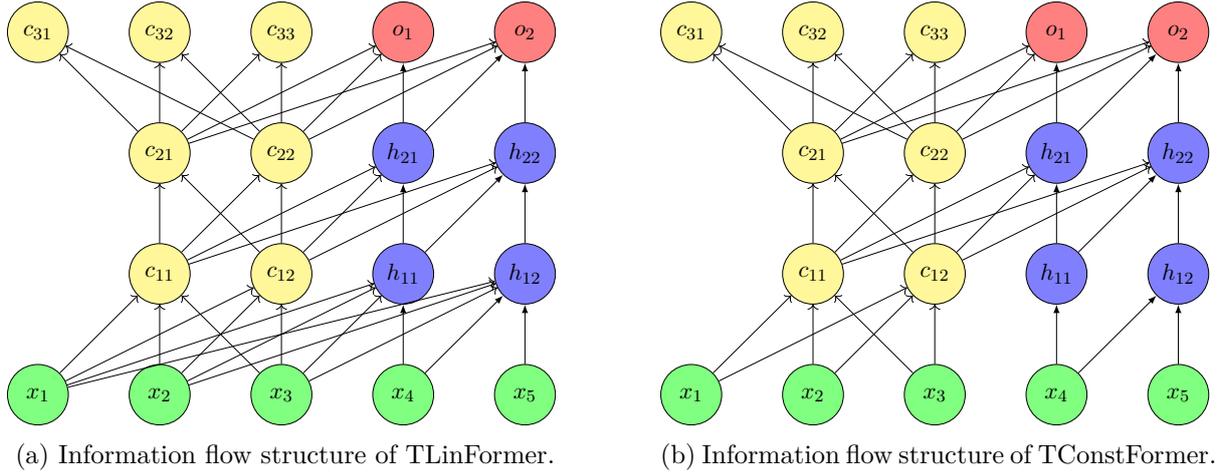

The core breakthrough of TConstFormer stems from a rethinking of this trade-off. We identified the key connections in TLinFormer responsible for long-term dependencies and selectively severed them. Specifically, as shown in Figure~\ref{subfig:TLinFormer_mlp_view}, we interrupted the direct information pathways from historical inputs (e.g., $x_1$ to $x_3$) to the current computational units (e.g., $h_{11}, h_{12}$). This structural modification makes the model's inference state no longer dependent on the entire growing historical sequence, but on a fixed-size hidden state, thereby fundamentally solving the problem of the KV Cache growing linearly with sequence length.

\section{TConstFormer Architecture}
\label{sec:architecture}

To implement the connection structure shown in Figure~\ref{subfig:TConstFormer_mlp_view}, we still use the same attention components as in TLinFormer, as illustrated in Figure~\ref{fig:attention_types}:
  \begin{figure}[H]
    \centering 

    \begin{subfigure}[b]{0.22\textwidth} 
        \centering
        \begin{tikzpicture}[
    neuron/.style={circle, draw, minimum size=0.8cm}, 
    input/.style={neuron, fill=green!50},
    hidden/.style={neuron, fill=blue!50},
    output/.style={neuron, fill=red!50},
    >=latex,
    scale=0.8, transform shape 
]
    \def\numI{3}
    \def\numO{3}
    \genNodeLayer{I}{0}{0}{\numI}{input}{x}{}
    \genNodeLayer{O}{0}{3}{\numO}{output}{o}{}
    \genFullconnect{I}{1}{\numI}{O}{1}{\numO}
\end{tikzpicture}
        \caption{Self Attention}
        \label{subfig:self_attention}
    \end{subfigure}
    \hfill 
    \begin{subfigure}[b]{0.22\textwidth}
        \centering
        \begin{tikzpicture}[
    neuron/.style={circle, draw, minimum size=0.8cm},
    input/.style={neuron, fill=green!50},
    hidden/.style={neuron, fill=blue!50},
    output/.style={neuron, fill=red!50},
    >=latex,
    scale=0.8, transform shape
]
    \def\numI{3}
    \def\numO{3}
    \genNodeLayer{I}{0}{0}{\numI}{input}{x}{}
    \genNodeLayer{O}{0}{3}{\numO}{output}{o}{}
    \genCausalFullconnect{I}{1}{\numI}{O}{1}{\numO}
\end{tikzpicture}
        \caption{Causal Attention}
        \label{subfig:causal_attention}
    \end{subfigure}
    \hfill
    \begin{subfigure}[b]{0.22\textwidth}
        \centering
        \begin{tikzpicture}[
    neuron/.style={circle, draw, minimum size=0.8cm},
    input/.style={neuron, fill=green!50},
    hidden/.style={neuron, fill=blue!50},
    output/.style={neuron, fill=red!50},
    >=latex,
    scale=0.8, transform shape
]
    \def\numI{3}
    \def\numO{2}
    \genNodeLayer{I}{0}{0}{\numI}{input}{x}{}
    \genNodeLayer{O}{0.5}{3}{\numO}{output}{o}{}
    \genFullconnect{I}{1}{\numI}{O}{1}{\numO}
\end{tikzpicture}
        \caption{Focused Attention}
        \label{subfig:focused_attention}
    \end{subfigure}
    \hfill
    \begin{subfigure}[b]{0.22\textwidth}
        \centering
        \begin{tikzpicture}[
    neuron/.style={circle, draw, minimum size=0.8cm},
    input/.style={neuron, fill=green!50},
    hidden/.style={neuron, fill=blue!50},
    output/.style={neuron, fill=red!50},
    context/.style={neuron, fill=yellow!50}, 
    >=latex,
    scale=0.8, transform shape
]
    \def\numC{1}
    \def\numI{2}
    \def\numO{2}

    \genNodeLayer{C}{0}{0}{\numC}{context}{c}{}
    \genNodeLayer{I}{1}{0}{\numI}{input}{x}{}
    \genNodeLayer{O}{0.5}{3}{\numO}{output}{o}{}
    \genFullconnect{C}{1}{\numC}{O}{1}{\numO}
    \genFullconnect{I}{1}{\numI}{O}{1}{\numO}
\end{tikzpicture}
        \caption{Cross Attention}
        \label{subfig:cross_attention}
    \end{subfigure}
    
    \caption{Connection diagrams for the 4 types of attention mechanisms required in this paper.}
    \label{fig:attention_types}
\end{figure}
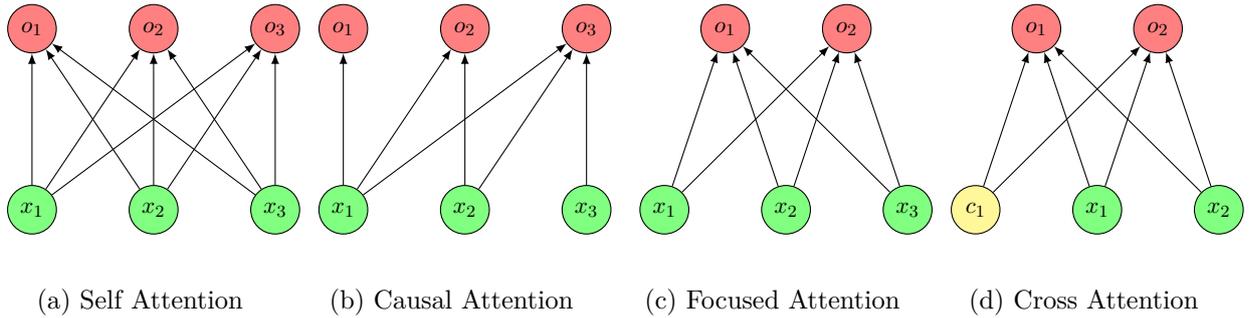

All the types above are special cases of the attention mechanism, calculated by the formula:
\[
\text{Attention}(\bm{Q}, \bm{K}, \bm{V}) = \text{softmax}\left(\frac{\bm{Q} \bm{K^T}}{\sqrt{d_k}}\right) \bm{V}
\]
We no longer view it as the traditional "query-key-value" interaction, but rather examine the attention mechanism from the perspective of dimensionality transformation. Assuming Q has dimensions (B, $L_Q$, D), and K and V have dimensions (B, $L_K$, D), the result of attention has dimensions (B, $L_Q$, D). We can see this as the attention mechanism scaling the L dimension of K and V, i.e., scaling from the original $L_K$ to $L_Q$. From an information flow perspective, attention completes a thorough fusion of the information in Q, K, and V. It is this picture that allows us to understand attention computation as a type of fully connected layer acting on the L dimension (like an MLP for the L dimension). Based on this understanding, we can precisely construct the information flow required by Figure~\ref{subfig:TConstFormer_mlp_view} by designing and combining different patterns of attention.

A TConstFormer block operates on an input partitioned into a historical context window $X_{\mathit{hist}}$ and a generation window $X_{\mathit{gen}}$. Its topological connections are constructed layer by layer as follows:
\begin{enumerate}
    \item \textbf{Context Path Encoding}: The historical context $X_{\mathit{hist}}$ is compressed in the L dimension in the first layer using the attention shown in Figure~\ref{subfig:focused_attention}. Subsequent intermediate layers are processed by self-attention layers. The final layer restores the L dimension using the attention shown in Figure~\ref{subfig:cross_attention} (of course, if stacking multiple TConstFormer blocks is not considered, the computation of the final layer can be omitted).

    \item \textbf{Generation Path Computation}: At each layer $i$, the computation involves two information flows:
    \begin{itemize}
        \item \textbf{Internal Cohesion (Causal Self-Attention)}: A causal self-attention mechanism is applied to the generation window representation from the previous layer ($H_{i-1}$, where $H_0 = X_{\mathit{gen}}$). This allows tokens within the generation window to interact with each other while adhering to causal constraints.
        \item \textbf{Context Integration (Cross-Attention)}: Historical information is fused into the generation window using a cross-attention mechanism. The queries come from the generation path ($H_{i-1}$), while the keys and values come from ($C_{i-1}$).
        \item The results of these two attention mechanisms are combined and passed through a feed-forward network (FFN) to produce the output of the current layer $H_i$.
    \end{itemize}
\end{enumerate}

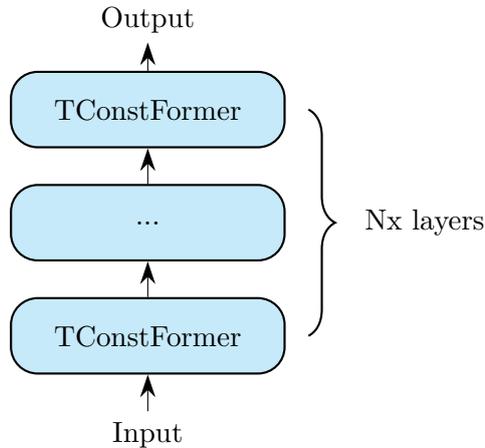
\begin{figure}[H]
    \centering
    \begin{tikzpicture}
    \pgfmathsetmacro {\height} {1}
    \pgfmathsetmacro {\gap} {0.5}
    \pgfmathsetmacro {\num} {3}
    \pgfmathsetmacro {\yposFinal} {\height * \num + (\num + 1) * \gap}

    \foreach \i in {1,...,\num} {
        \pgfmathsetmacro{\ypos}{(\height + \gap) * (\i - 1) + \height/2};

        \pgfmathparse{ifthenelse(\i==2, 1, 0)}
        \ifnum\pgfmathresult=1
            \def\mtext{...}       
        \else
            \def\mtext{TConstFormer}  
        \fi
        \node[draw, rounded corners=10pt, fill=cyan!20, thick, minimum width=3.6cm, minimum height=\height cm] (T\i) at (2,\ypos cm) {\mtext};
    }
 
    \foreach \i in {1,...,2} {
        \pgfmathtruncatemacro{\nexti}{\i + 1}   
        \draw[-{Stealth[length=3mm, width=2mm]}] (T\i.north)  -- (T\nexti.south) ;
    }
    
    \node[below] (TEXT-1) at (2,-\gap cm) {Input};
    \node[below] (TEXT-2) at (2,\yposFinal cm) {Output};

    \draw[-{Stealth[length=3mm, width=2mm]}] (TEXT-1.north)  -- (T1.south) ;
    \draw[-{Stealth[length=3mm, width=2mm]}] (T\num.north)  -- (TEXT-2.south) ;
    
    \draw[black,decorate,decoration={brace,amplitude=10},thick]
    ([xshift=0.3cm]T\num.east)  -- ([xshift=0.3cm]T1.east) node[midway, right=0.6cm] {Nx layers};

\end{tikzpicture}
    \caption{Schematic of a stacked TConstFormer network structure.}
    \label{fig:stacked_TConstFormer}
\end{figure}

A TConstFormer block can function as a standalone module or be stacked repeatedly to form a deep network. When used alone, the attention in the final layer of the historical window in Figure~\ref{subfig:TConstFormer_mlp_view} ($C_3$) can be omitted. When multiple blocks are stacked, as shown in Figure~\ref{fig:stacked_TConstFormer}, the output of each layer serves as the input for the next, thereby constructing a standard deep Transformer decoder structure.

\section{Attention Complexity Analysis}
\label{sec:complexity}

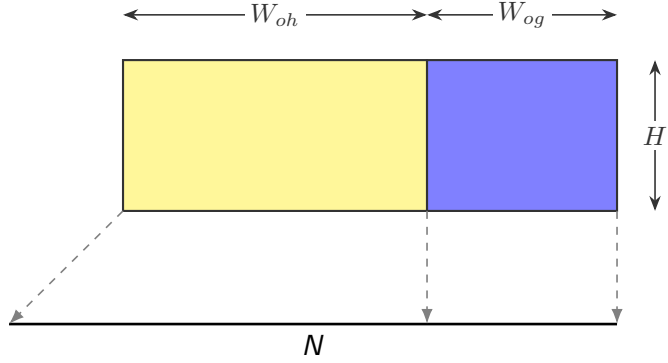
\begin{figure}[H]
    \centering
    \begin{tikzpicture}[
    font=\sansmath\sffamily, 
    rect_style/.style={draw=black!80, line width=0.8pt},             
    proj_line/.style={-Latex, line width=0.6pt, gray, dashed},      
    dim_arrow/.style={<->, >=Stealth, line width=0.6pt, black!80},   
    dim_label/.style={fill=white, inner sep=2pt, font=\small}        
]
    \colorlet{colorA}{yellow!50}
    \colorlet{colorB}{blue!50}

    \def\woh{4}
    \def\w{2.5}
    \def\h{2}
    \def\baselineY{-1.5}
    \def\dimlineY{2.6}

    \begin{scope}[blur shadow={shadow opacity=25, shadow blur steps=5}]
        \filldraw[fill=colorA, rect_style] (0,0) rectangle (\woh, \h);
        \filldraw[fill=colorB, rect_style] (\woh,0) rectangle (\woh + \w, \h);
    \end{scope}

    \draw[line width=1pt] (-1.5, \baselineY) -- (\woh + \w, \baselineY) 
        node[midway, below]{$N$};

    \draw[proj_line] (0,0) -- (-1.5, \baselineY);
    \draw[proj_line] (\woh,0) -- (\woh, \baselineY);
    \draw[proj_line] (\woh + \w, 0) -- (\woh + \w, \baselineY);

    \draw[dim_arrow] (0, \dimlineY) -- (\woh, \dimlineY)
        node[midway, dim_label] {$W_{oh}$};
    \draw[dim_arrow] (\woh, \dimlineY) -- (\woh + \w, \dimlineY)
        node[midway, dim_label] {$W_{og}$};

    \pgfmathsetmacro {\startXh} {\woh + \w + 0.5}
    \draw[dim_arrow] (\startXh, 0) -- (\startXh, \h)
        node[midway, dim_label] {$H$};

\end{tikzpicture}
    \caption{Windowed computation schematic.}
    \label{fig:windowed_div}
\end{figure}

Let the total input sequence length be $N$, the feature dimension be $D$, and the number of intermediate self-attention layers within a TConstFormer block be $H$. The model is partitioned into a window of length $\Woh$ for processing historical context, which can observe a historical sequence of length $N - \Wog$, and a generation window of length $\Wog$.

A detailed derivation of the following discussion can be found in Appendix~\ref{app:complexity_derivation}.

\subsection{Cache Miss}

A Cache Miss refers to an event where pre-computed results cannot be reused, requiring a full computation from scratch. It primarily occurs in two scenarios:
\begin{enumerate}
\item \textbf{Training Phase}: Since the historical context is different for each training batch, the caching mechanism is not applicable, and every forward pass can be considered a Cache Miss.
\item \textbf{Inference Phase}:
\begin{itemize}
\item When generating the initial token for a given context.
\item At the moment of generating the first new token after the historical context window is updated (i.e., slid).
\end{itemize}
\end{enumerate}

The computational cost of a cache miss is equivalent to performing one full forward computation with the cache disabled.

The total computational cost is the sum of the costs for the context window and the generation window. Crucially, the total cost is a precise linear function of the total sequence length, of the form:

\begin{equation}
\text{Total Cost} = C_1 \cdot N + C_0
\label{eq:total_cost_simple_cache_miss}
\end{equation}
where the slope and intercept are constants determined by the model's hyperparameters:
\begin{align}
C_1 &= D \cdot (2\Woh) \\
C_0 &= D \left[ H(\Woh^2 + \Wog^2 + \Wog\Woh) + 2\Wog^2 - \Wog\Woh \right] \label{eq:intercept_c0_updated}
\end{align}

\begin{equation}
\text{Total Cost} = D \left[ N(2\Woh) + H(\Woh^2 + \Wog^2 + \Wog\Woh) + 2\Wog^2 - \Wog\Woh \right]
\label{eq:total_cost_full}
\end{equation}

Since $D, \Woh, \Wog$, and $H$ are fixed or bounded after training, \textbf{the computational complexity of TConstFormer is strictly linear with respect to the sequence length $N$}.

\subsection{Cache Hit}

A Cache Hit occurs only during autoregressive inference. It refers to the event of generating any subsequent token other than the first one within a single generation cycle (i.e., when the historical context window remains static).

\textbf{The total computational cost is a constant quantity:}
\begin{equation}
\text{Total Cost} = (H + 1) D \Woh + (H+2)D \Wog^2
\label{eq:total_cost_simple_cache_hit}
\end{equation}

\subsection{Complexity Summary}

The computational complexity of TConstFormer exhibits a dual-mode characteristic closely tied to the cache state, which is the core of its efficient inference capability.

\begin{itemize}
    \item \textbf{On a Cache Miss}, such as during training or when generating the initial token, the model's computational cost is \textbf{strictly linear} with the total sequence length $N$, with a complexity of $\mathcal{O}(N)$. This constitutes the upper bound of the model's single computation overhead.

    \item \textbf{On a Cache Hit}, i.e., when generating subsequent tokens in autoregressive inference, the computational cost is \textbf{completely independent} of the total sequence length $N$, becoming a \textbf{constant} determined only by the window sizes ($\Woh, \Wog$) and model depth ($H$), with a complexity of $\mathcal{O}(1)$.
\end{itemize}

This dynamic transition from linear to constant cost allows TConstFormer to maintain the overhead of the vast majority of generation steps at an extremely low level when processing long sequences, thereby achieving orders-of-magnitude inference acceleration.

\section{Model Properties and Discussion}
\label{sec:discussion}

\subsection{Training Process}
\label{sec:discussion_of_windows}

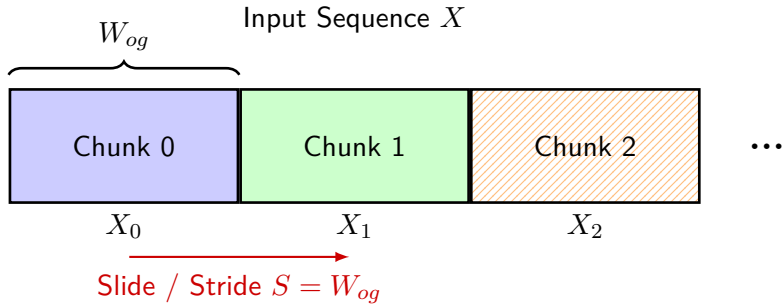
\begin{figure}[H]
    \centering
    \begin{tikzpicture}[
        font=\sffamily,
        >=latex,
        block/.style={draw, rectangle, minimum height=1.5cm, minimum width=3cm, line width=1pt},
        brace/.style={decorate, decoration={brace, amplitude=5pt, raise=5pt}, line width=1pt}
    ]

    \node[block, fill=blue!20, label=below:\(X_0\)] (block0) {Chunk 0};

    \node[block, fill=green!20, label=below:\(X_1\), right=0cm of block0] (block1) {Chunk 1};

    \node[block, fill=orange!20, pattern=north east lines, pattern color=orange!50, label=below:\(X_2\), right=0cm of block1] (block2) {Chunk 2};

    \node[right=0.5cm of block2, font=\Large] (ellipsis) {\textbf{...}};
    
    \node[above=0.6cm of block1] {Input Sequence \(X\)};

    \draw[brace] (block0.north west) -- (block0.north east) node[midway, above=10pt] {$W_{og}$};
    
    \draw[->, thick, red!80!black, shorten <=2pt, shorten >=2pt] 
        ([yshift=-20pt]block0.south) -- 
        node[midway, below=2pt, text=red!80!black] {Slide / Stride \(S = W_{og}\)} 
        ([yshift=-20pt]block1.south);

    \end{tikzpicture}
    
    \caption{Sliding window information processing flow during training.}
    \label{fig:wog_slide}
\end{figure}

The model processes long sequences in chunks. During training, it employs a sliding window mechanism. First, it processes the interval $[0, \Wog]$, where the historical context is empty. Then, the window slides by a distance of $\Wog$, and the model processes the interval $[\Wog, 2\Wog]$, using the first chunk $[0, \Wog]$ as its historical context. The window then slides again by $\Wog$, and the model processes the interval $[2\Wog, 3\Wog]$, using the first two chunks $[0, 2\Wog]$ as its historical context. This process continues until the entire sequence has been processed. The outputs of each generation window are concatenated to form the final output sequence for loss calculation.

\subsection{Excellent Cache Efficiency during Inference}
\label{sec:cache_efficiency}

\subsubsection{$\mathcal{O}(1)$ KV Cache Footprint}
The KV cache of a standard Transformer scales linearly with the entire sequence length $L$, as shown in Equation~\eqref{eq:kv_cache_memory}, which often becomes a memory bottleneck.
\begin{equation}
M_{\text{transformer}} = 2  \cdot B \cdot L \cdot d_{\text{model}} \cdot P_{\text{bytes}} \cdot N_{\text{layers}}
\label{eq:kv_cache_memory}
\end{equation}

Here, $N_{\text{layers}}$ is the number of layers, $B$ is the batch size, $d_{\text{model}}$ is the hidden dimension, and $P_{\text{bytes}}$ is the precision in bytes.

TConstFormer offers a significant advantage in autoregressive inference. The key-value (KV) cache associated with the historical context window ($\Woh$) remains static as long as the context itself does not change. A cache invalidation and re-computation event only occurs after $\Wog$ new tokens have been generated and the context window needs to be updated.

The cache overhead of TConstFormer comes from the KV Cache consumption within the historical and generation windows. The final $M_{\text{TConstFormer}}$ is:
\begin{equation}
M_{\text{TConstFormer}}  = 2 \cdot B \cdot (H + 1) \cdot \Woh \cdot d_{\text{model}} +  2 \cdot B \cdot (H + 2) \cdot \Wog \cdot d_{\text{model}}
\label{eq:tlin_cache_approx}
\end{equation}

The KV Cache footprint of TConstFormer is \textbf{constant and completely decoupled from the sequence length}, achieving a true ${\mathcal{O}}(1)$ space complexity.

\subsubsection{Superior Time Acceleration}

Standard autoregressive Transformers (Decoder-only) face a fundamental efficiency bottleneck during inference. In contrast, as discussed in Section~\ref{sec:complexity}, when a cache hit occurs, TConstFormer's inference time is constant.

\section{Experiments}
\label{sec:experiments}

In this section, we will validate the effectiveness of TConstFormer through a series of experiments. We begin by detailing the experimental setup, including the baselines, model configurations, and evaluation metrics, to ensure fairness and reproducibility. Subsequently, we will present and thoroughly analyze the main results on the wikitext-103-v1 benchmark.

\subsection{A Note on Long-Context Retrieval Tasks}
The "Needle in a Haystack" benchmark is widely used to evaluate the long-context retrieval capabilities of Large Language Models (LLMs). However, this test primarily measures the complex instruction-following and long-range dependency abilities that emerge after pre-training on massive, diverse corpora. The TConstFormer architecture proposed in this paper focuses its core contribution on demonstrating a fundamental improvement in computational and memory efficiency. Due to the limitations of our model scale (41M parameters) and training data (Wikipedia), its design objective is not to replicate the full range of emergent abilities of large-scale LLMs. Therefore, we consider the "Needle in a Haystack" test to be orthogonal to the core goal of validating architectural efficiency that this paper aims to address, and thus have not included it in our primary evaluation scope. Extending the TConstFormer architecture to larger-scale models to explore its potential in complex retrieval tasks is a promising direction for future research.

\subsection{Implementation Details}
\label{ssec:implementation_details}

\paragraph{Hardware Environment:}
All our training, inference, and testing were conducted on a platform more aligned with consumer-grade hardware. The platform is configured as follows: one \textbf{NVIDIA GeForce RTX 4090 GPU} (24 GB VRAM), one \textbf{AMD EPYC 7543 CPU}, and 62 GB of system memory.

\paragraph{Software Stack:}
The software stack was consistent across environments: \textbf{Python 3.12.11}, \textbf{PyTorch 2.7.1}, \textbf{CUDA 12.6}, \textbf{cuDNN 9.5.1}, \textbf{Hugging Face Transformers (v4.55.2)}, and the operating system was \textbf{Ubuntu 22.04.5 LTS}.

\subsubsection{Principle of Fair Comparison and Model Configuration}
\label{sssec:fair_comparison_and_config}

To ensure a fair and meaningful comparison between TConstFormer and the standard Transformer baseline, all experiments adhere to the principle of \textbf{parameter parity}. The core innovation of TConstFormer lies in the \textbf{reorganization} of information flow, rather than the introduction of new parameterized components. It is essentially a \textbf{topological reconstruction} of standard Transformer modules. Therefore, as long as the total computational depth of a stacked TConstFormer model matches the number of layers in a standard Transformer model, their \textbf{total parameter counts are identical}. This allows us to attribute any performance differences solely to the superiority of the architectural design.

In these experiments, we use a small-scale model configuration with approximately \textbf{41M} parameters. Both the baseline model and our TConstFormer use the same core hyperparameters:
\begin{itemize}
    \item \texttt{vocab\_size}: 50257 (same as GPT-2)
    \item \texttt{n\_embd} (embedding dimension): 432
    \item \texttt{n\_head} (number of attention heads): 12
    \item \texttt{n\_transformer\_block} (equivalent total depth): 8
\end{itemize}
For the baseline model, this is a standard 8-layer decoder-only Transformer. For TConstFormer, this equivalent depth of 8 is achieved by stacking 2 TConstFormer blocks, with each block having an internal depth hyperparameter of $H=2$.

Training parameters, such as learning rate, were kept identical for all models. The equivalent batch size was set to 256 (achieved through gradient accumulation).

\paragraph{A Note on Hyperparameter Selection:}
The choice of core hyperparameters, such as the internal depth $H=2$, was primarily guided by a pragmatic balance between computational overhead and the model's effective receptive field under limited computing resources. We believe that a comprehensive hyperparameter search tailored to different model scales is an essential step for future work, but it falls beyond the scope of this initial validation study.

\subsubsection{Dataset and Evaluation Metrics}
\label{sssec:dataset_and_metrics}

We use the wikitext-103-v1 dataset from the Hugging Face repository (Salesforce/wikitext) for all experiments. This dataset contains approximately 120 million tokens. Model performance is evaluated using Perplexity (PPL) on the validation set, where lower values indicate better performance.

\subsubsection{Baselines and Model Variants}
\label{sssec:baselines}

We compare TConstFormer against a standard decoder-only Transformer baseline and TLinFormer. To evaluate performance under different configurations, we trained multiple variants for each architecture. The naming convention for these variants is explained below:

\begin{description}
    \item[\texttt{Base XXX:}] 
    Represents the standard Transformer baseline. The suffix \texttt{XXX} denotes the sequence length used during its training. For example, \texttt{Base 1K} refers to the baseline model trained with a 1K sequence length.

    \item[\texttt{TConstFormer/TLinFormer XXX-YYY-ZZZ:}]
    Represents our TConstFormer/TLinFormer models, with the name composed of three parameters:
    \begin{itemize}
        \item \texttt{XXX}: The total sequence length used during training.
        \item \texttt{YYY}: The total length of the core observation window, i.e., $W_{total} = \Woh + \Wog$.
        \item \texttt{ZZZ}: The ratio of the historical context observation window to the total observation window, i.e., $\Woh/W_{total}$.
    \end{itemize}
    For example, \texttt{TConstFormer 2K-512-0.5} represents a TConstFormer model trained with a 2K sequence length, a total observation window of 512, where the historical context window is set to half the total window size ($0.5 \times 512 = 256$).
\end{description}

\subsection{Training Results and Analysis}
\label{ssec:results}

\subsubsection{Analysis of Training Overhead and Trade-offs}

\begin{figure}[H]
    \centering

    \begin{subfigure}[b]{0.3\textwidth}
        \centering
        \includegraphics[width=\linewidth]{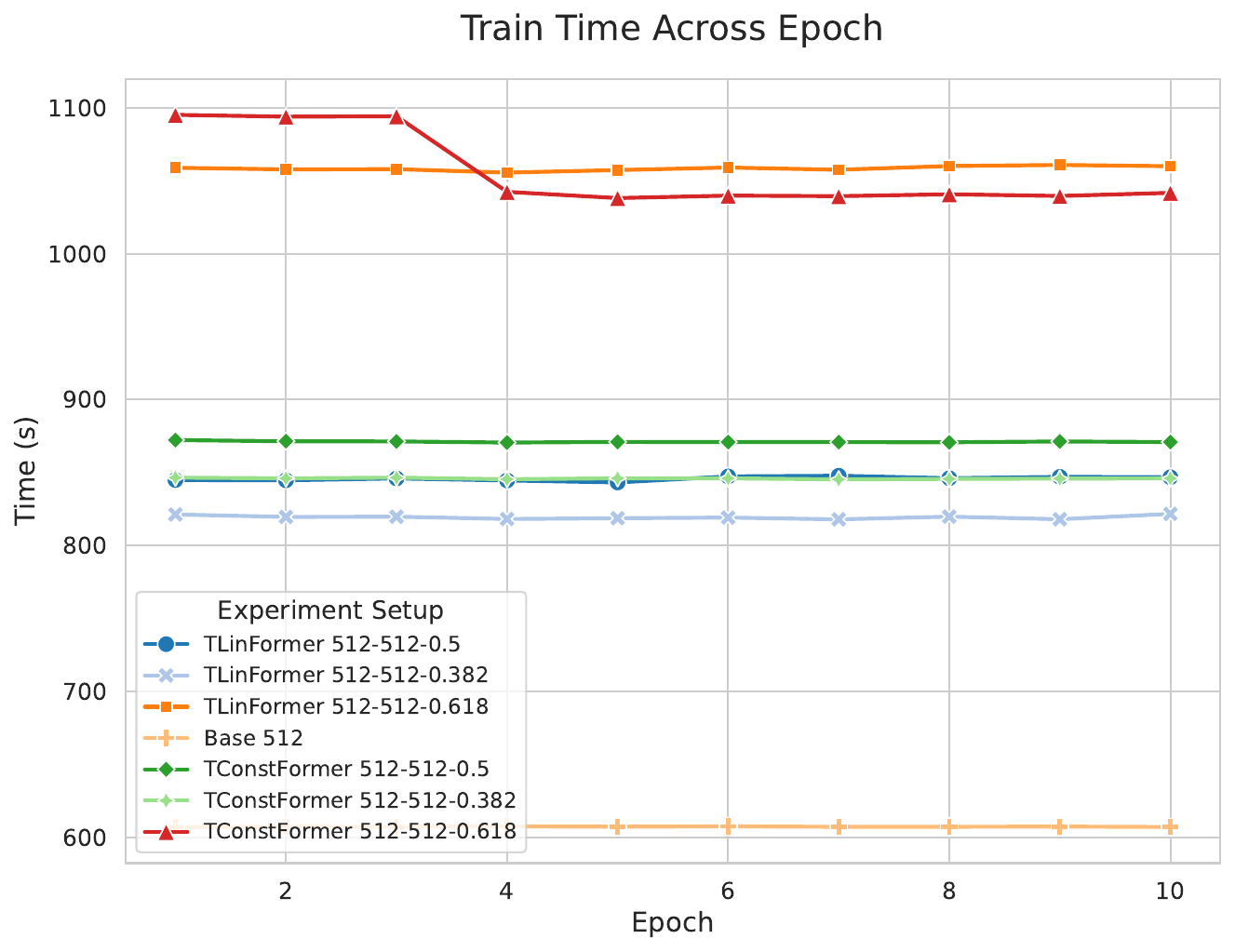}
        \caption{Training time per epoch for sequence length 512}
        \label{subfig:train_time_512}
    \end{subfigure}%
    \hfill
    \begin{subfigure}[b]{0.3\textwidth}
        \centering
        \includegraphics[width=\linewidth]{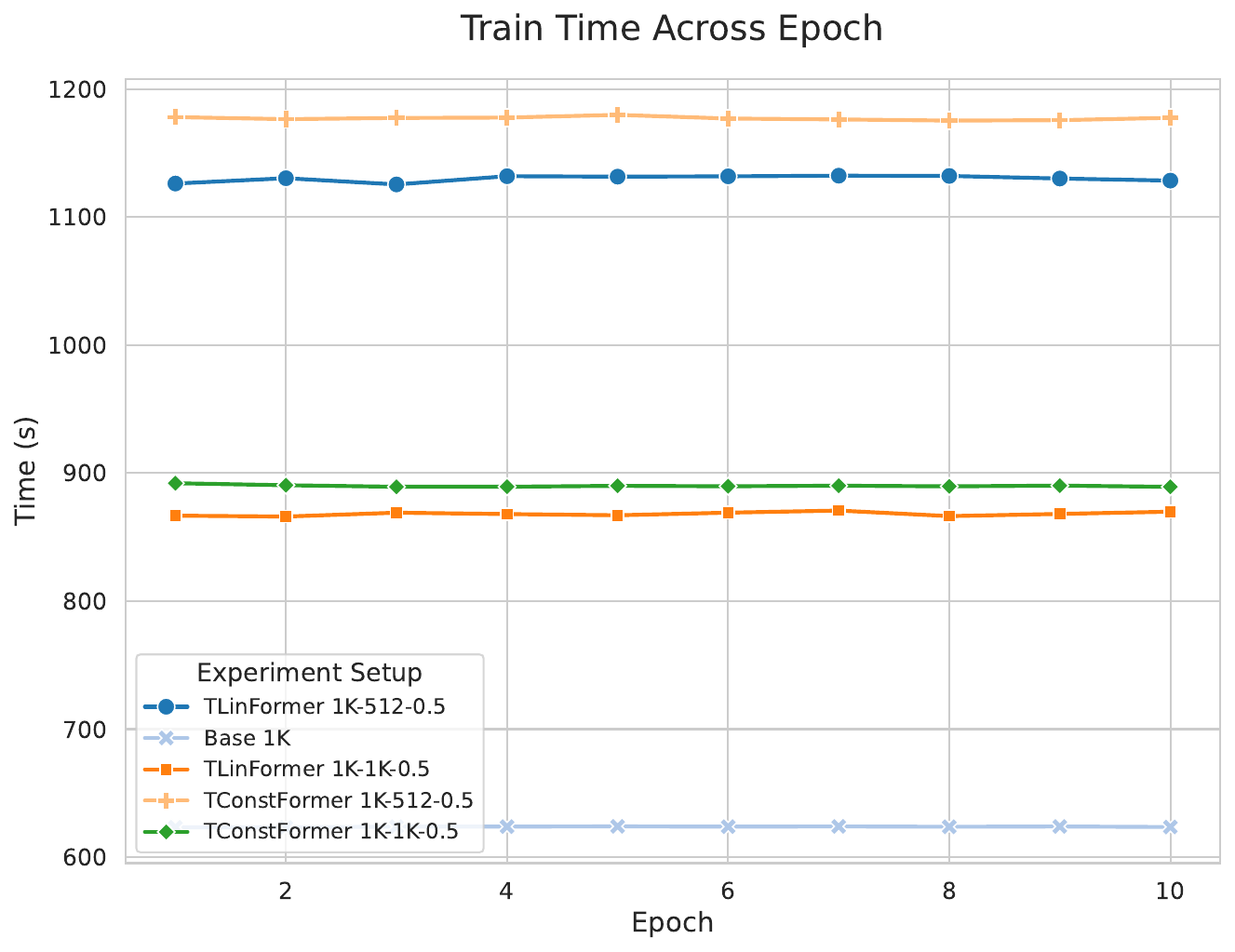}
        \caption{Training time per epoch for sequence length 1K}
        \label{subfig:train_time_1k}
    \end{subfigure}
    \hfill
    \begin{subfigure}[b]{0.3\textwidth}
        \centering
        \includegraphics[width=\linewidth]{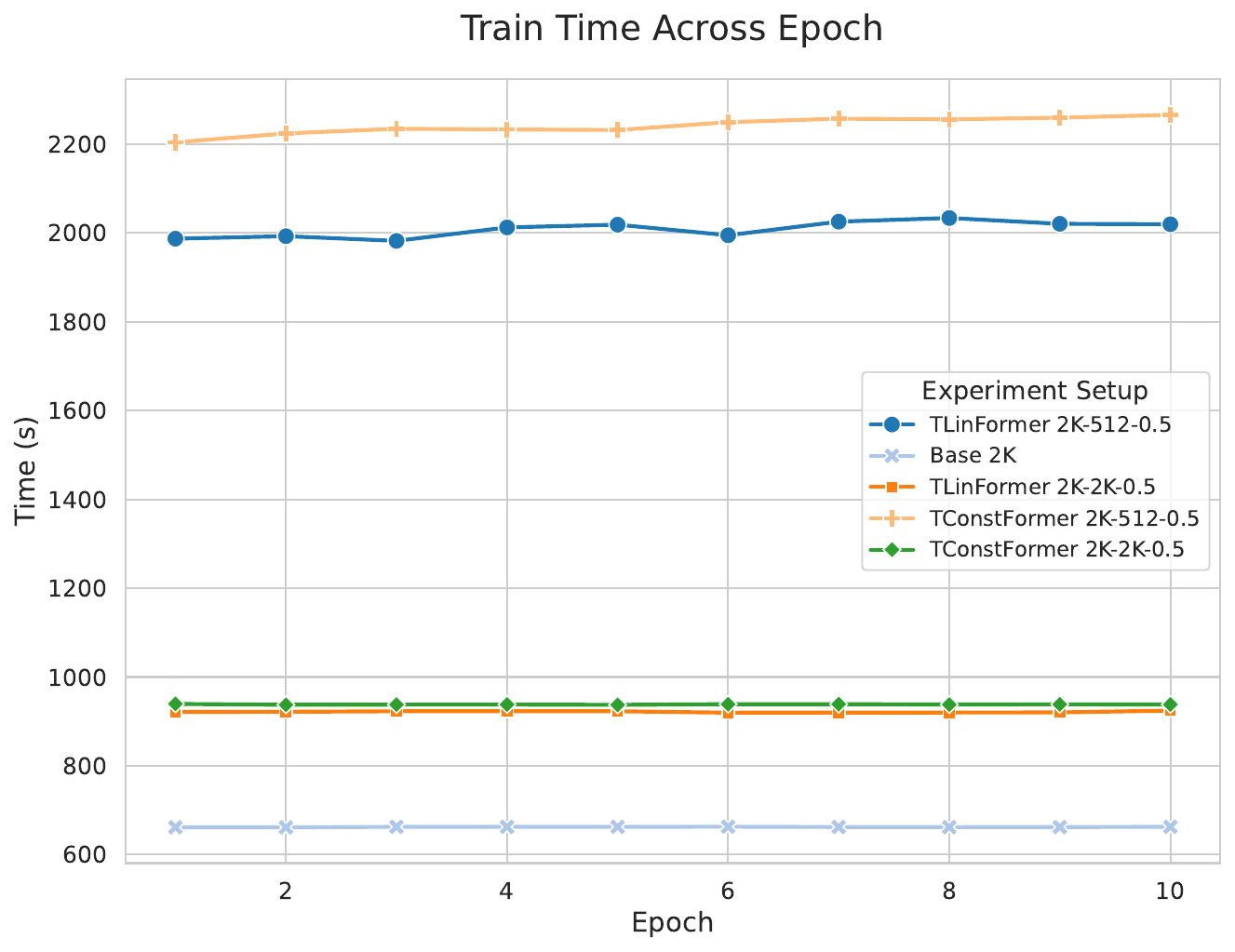}
        \caption{Training time per epoch for sequence length 2K}
        \label{subfig:train_time_2k}
    \end{subfigure}

    \caption{
        \textbf{Training efficiency comparison at different sequence lengths.}
        The plots show the wall-clock time required for each model to complete a single epoch at training sequence lengths of 512, 1K, and 2K.
    }
    \label{fig:train_time}
\end{figure}

We further evaluated the training efficiency of each architecture, as shown in Figure~\ref{fig:train_time}. It is important to note that to maximize hardware utilization across different sequence lengths, we adjusted the actual batch size and used gradient accumulation to ensure an equivalent batch size of 256 for all experiments. Therefore, a direct comparison of wall-clock time across different sequence lengths is not meaningful; we primarily focus on the relative efficiency at the same sequence length.

The results show that, under the same sequence length configuration, our new architectures exhibit a certain increase in training overhead compared to the baseline. This phenomenon is entirely consistent with our architectural design. For the \texttt{1K} sequence length, for instance, the baseline model processes the full 1K context in a single parallel pass. In contrast, our models (e.g., \texttt{TConstFormer 1K-1K-0.5}) use a chunked processing mechanism, for example, processing the first 512 tokens, then the next 512, and finally combining the results for loss calculation. While this chunked computation introduces additional scheduling overhead, it is precisely this design that forms the foundation for the model's efficient caching and significant performance gains during inference.

\paragraph{Quantitative Analysis of Training Overhead: }
To quantify this overhead more concretely, we analyzed the training time per epoch at a 1K sequence length. The baseline model, \texttt{Base 1K}, completed an epoch in approximately \textbf{620 seconds}, while our \texttt{TConstFormer 1K-1K-0.5} model took around \textbf{890 seconds}, representing an overhead of approximately \textbf{42\%}. We argue that this \textbf{controllable, one-time training cost} is a highly valuable and pragmatic engineering trade-off for achieving \textbf{orders-of-magnitude acceleration across countless inference tasks} throughout the model's lifecycle.

\subsubsection{Validation Set PPL}

\begin{figure}[H]
    \centering
    \includegraphics[width=0.8\textwidth]{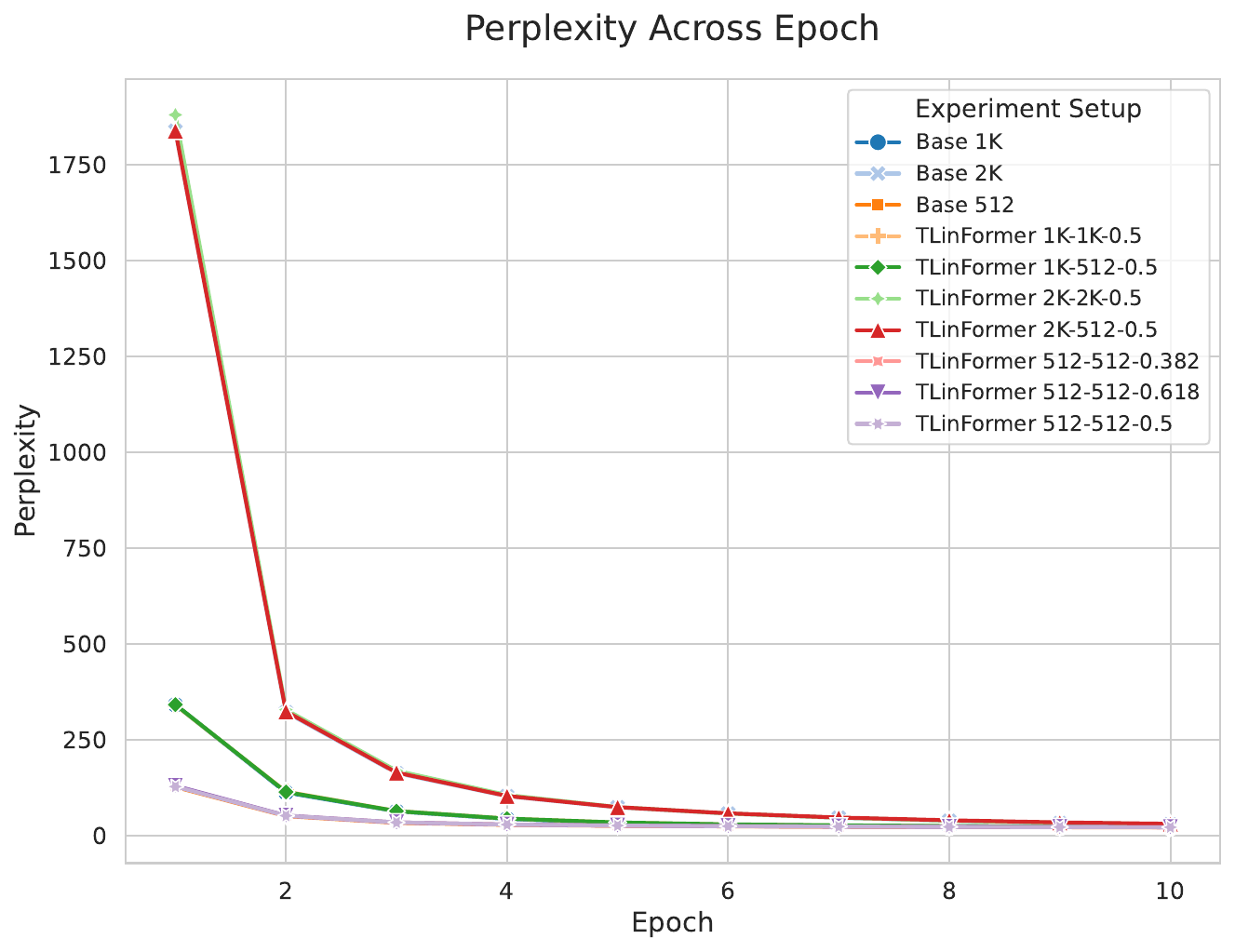} 
    \caption{Perplexity (PPL) of each model over training epochs.}
    \label{fig:ppl_of_tests}
\end{figure}

Table~\ref{tab:ppl_across_epochs} and Figure~\ref{fig:ppl_of_tests} show the perplexity of all model variants on the wikitext-103-v1 validation set as training progresses. We can observe several key findings from the results:

\begin{enumerate}
    \item \textbf{Architectural reconstruction does not sacrifice base performance.}
    First, we verified that our proposed architectural reconstruction does not introduce significant performance loss under equivalent configurations. As shown in Table~\ref{tab:ppl_across_epochs}, when the observation window length is identical to the baseline model's context length (e.g., comparing \texttt{Base 1K} with \texttt{TLinFormer 1K-1K-0.5} and \texttt{TConstFormer 1K-1K-0.5}), the final perplexities (PPL) are almost identical (22.5, 22.7, and 22.7, respectively). This strongly demonstrates that our proposed information flow reorganization is achieved without compromising the model's fundamental expressive power.

    \item \textbf{TConstFormer shows a performance advantage in equivalent configurations.}
    A key finding is that under identical training length and observation window configurations, TConstFormer consistently matches or outperforms TLinFormer, and generally reaches the performance of the baseline model. The most notable example is in the \texttt{512} context, where \texttt{TConstFormer 512-512-0.5} achieves a final PPL of \textbf{21.6}, matching \texttt{Base 512} and outperforming all TLinFormer variants (21.9). This indicates that TConstFormer's architectural optimization not only brings a leap in inference efficiency but also shows some advantage in model performance.

    \item \textbf{Controllable performance trade-off from information compression.}
    We also investigated the effect of "forced compression" (i.e., using an observation window shorter than the total sequence length) on performance. The results show that both TLinFormer and TConstFormer exhibit a slight and expected performance gap compared to the full-size baseline when processing long sequences with short observation windows (e.g., comparing \texttt{Base 1K} vs. \texttt{1K-512-0.5}). We believe this performance trade-off is a direct manifestation of the model being forced to compress and abstract long historical information within a limited window, which is a key step towards higher efficiency and stronger generalization.

    \item \textbf{High robustness to core hyperparameters.}
    Finally, we conducted an ablation study on the model's performance with different historical window ratios ($\Woh/W_{\text{total}}$). In the \texttt{512-512-X} configuration group, for both TLinFormer and TConstFormer, despite the hyperparameter changing from 0.382 to 0.618, the final PPL of all variants remained stable within a very small range. This proves that our architecture's performance advantage stems from its robust core design, rather than fine-tuning of specific hyperparameters, greatly enhancing its reliability and ease of use in practical applications.
\end{enumerate}

\paragraph{Unexpected Performance Improvement from Architectural Simplification:} \mbox{}

A noteworthy and counter-intuitive finding is that TConstFormer exhibited performance that matched or even surpassed TLinFormer in several configurations (see Table~\ref{tab:ppl_across_epochs}). We simplified TLinFormer architecturally by removing the direct connections between the generation window and the full historical sequence. Our initial intuition was that more information pathways should lead to greater expressive power. However, the experimental results suggest that this simplification turned out to be an advantage.

We speculate that this performance improvement stems from a stronger \textbf{Structured Inductive Bias} introduced by TConstFormer, which forces the network to achieve a clearer \textbf{Functional Specialization}:
\begin{itemize}
    \item The \textbf{historical context window} is specifically shaped into an efficient \textbf{information encoding module}. Its sole responsibility is to distill the ever-growing sequence history into a bounded-size, high-information-density representation of the "world state."
    \item The \textbf{generation window}, in turn, focuses on its core task as a \textbf{language generation module}. It no longer needs to weigh between raw, unrefined historical details and a compressed summary, but can make decisions based on a more stable, abstract, and high-quality source of information.
\end{itemize}
We believe that this clear modular division of labor may simplify the model's optimization process and guide it to learn a more robust and generalizable internal representation.

\begin{table}[H]
  \centering
  \caption{Perplexity (PPL) on the wikitext-103-v1 validation set. Lower is better.}
  \label{tab:ppl_across_epochs}
  \sisetup{table-format=4.1}
  \resizebox{\textwidth}{!}{
    \begin{tabular}{@{}l S S S S S S S S S S@{}}
        \toprule
        \multirow{2}{*}{\textbf{experiment}} & \multicolumn{10}{c}{\textbf{Epoch}} \\
        \cmidrule(l){2-11}
        & {1} & {2} & {3} & {4} & {5} & {6} & {7} & {8} & {9} & {10} \\
        \midrule
        Base 512	&	126.8 	&	51.0 	&	33.4 	&	28.1 	&	25.7 	&	24.2 	&	23.3 	&	22.6 	&	22.0 	&	21.6 	\\
        TLinFormer 512-512-0.382	&	129.1 	&	51.9 	&	34.4 	&	28.7 	&	26.2 	&	24.6 	&	23.6 	&	22.8 	&	22.4 	&	21.9 	\\
        TLinFormer 512-512-0.5	&	127.9 	&	52.1 	&	34.5 	&	28.9 	&	26.3 	&	24.7 	&	23.7 	&	22.9 	&	22.3 	&	21.9 	\\
        TLinFormer 512-512-0.618	&	130.1 	&	52.4 	&	34.4 	&	28.8 	&	26.2 	&	24.8 	&	23.6 	&	23.0 	&	22.4 	&	21.9 	\\
        TConstFormer 512-512-0.382	&	127.3 	&	51.5 	&	34.1 	&	28.6 	&	26.1 	&	24.5 	&	23.6 	&	22.8 	&	22.2 	&	21.8 	\\
        TConstFormer 512-512-0.5	&	124.3 	&	51.0 	&	33.7 	&	28.2 	&	25.7 	&	24.1 	&	23.2 	&	22.5 	&	22.0 	&	21.6 	\\
        TConstFormer 512-512-0.618	&	129.4 	&	52.0 	&	34.4 	&	28.9 	&	26.2 	&	24.8 	&	23.7 	&	23.0 	&	22.5 	&	22.0 	\\      
        \midrule
        Base 1K	&	341.5 	&	112.5 	&	63.4 	&	43.5 	&	33.0 	&	28.3 	&	25.8 	&	24.2 	&	23.3 	&	22.5 	\\
        TLinFormer 1K-1K-0.5	&	340.7 	&	115.0 	&	64.3 	&	44.6 	&	34.2 	&	29.0 	&	26.3 	&	24.6 	&	23.5 	&	22.7 	\\
        TLinFormer 1K-512-0.5	&	341.8 	&	114.0 	&	64.0 	&	44.5 	&	34.3 	&	29.4 	&	26.8 	&	25.0 	&	23.8 	&	23.0 	\\
        TConstFormer 1K-1K-0.5	&	343.2 	&	114.0 	&	63.8 	&	44.5 	&	33.8 	&	28.8 	&	26.2 	&	24.6 	&	23.5 	&	22.7 	\\
        TConstFormer 1K-512-0.5	&	342.1 	&	113.6 	&	64.5 	&	44.5 	&	34.2 	&	29.3 	&	26.7 	&	24.9 	&	23.9 	&	23.0 	\\
        \midrule
        Base 2K	&	1839.4 	&	321.8 	&	162.9 	&	102.9 	&	73.9 	&	57.5 	&	46.6 	&	38.8 	&	32.7 	&	29.5 	\\
        TLinFormer 2K-2K-0.5	&	1881.0 	&	328.5 	&	169.4 	&	106.0 	&	74.6 	&	57.7 	&	46.8 	&	38.9 	&	33.7 	&	29.8 	\\
        TLinFormer 2K-512-0.5	&	1839.7 	&	324.0 	&	164.6 	&	103.5 	&	74.3 	&	58.2 	&	47.2 	&	40.1 	&	34.5 	&	30.9 	\\
        TConstFormer 2K-2K-0.5	&	1945.7 	&	327.1 	&	167.7 	&	104.3 	&	74.4 	&	57.2 	&	46.4 	&	38.8 	&	33.3 	&	29.6 	\\
        TConstFormer 2K-512-0.5	&	1852.0 	&	326.5 	&	165.7 	&	103.9 	&	74.5 	&	58.4 	&	46.9 	&	39.4 	&	33.9 	&	30.3 	\\      
        \bottomrule
    \end{tabular}
  }
\end{table}

\subsection{Inference Results and Analysis}

\subsubsection{Testing Methodology}
\begin{enumerate}
    \item \textbf{Precondition:} Caching is always enabled.
    \item \textbf{Environment Initialization:} Before each test run, we clear the GPU memory by calling \texttt{torch.cuda.empty\_cache()} to ensure each test starts from a clean, consistent initial state, eliminating interference from caching.
    \item \textbf{Incremental Sequence Length:} We start with a small initial sequence length (e.g., $N=1$) and then incrementally increase the sequence length $N$ by a fixed step (e.g., 10,000 tokens).
    \item \textbf{Inference and Timing:} For each initial sequence length $N$, we generate a random integer tensor of shape $(1, N)$ as input. This tensor is fed into the model to generate 6 new tokens. We measure and record the time and cache consumption required to generate each token. We always select the first (cache miss, equivalent to cache off) and third token (cache hit) for detailed analysis.
    \item \textbf{Determining Maximum Sequence Length:} After clearing the model's cache, we continuously increase $N$ and repeat step 3 until the model fails to complete inference due to an Out of Memory (OOM) error.
\end{enumerate}

\subsubsection{Inference Time Complexity Analysis (Figures a, b, c)}

\paragraph{The Baseline Model's Bottleneck:}
As shown in Figure~\ref{fig:inference_analysis}(a), the inference latency of the standard Transformer baseline exhibits a near-quadratic $\mathcal{O}(N^2)$ trend with increasing sequence length $N$. This phenomenon reveals a gap between theory and practice: although the KV cache reduces the theoretical computational complexity to $\mathcal{O}(N)$, in practice, the performance bottleneck shifts from floating-point operations (FLOPs) to \textbf{Memory IO}, dominated by memory copy operations.

\noindent\textit{A note on pre-allocation strategies:} While engineering tricks like pre-allocating a larger memory space can mitigate this issue, it fundamentally comes at the cost of higher static memory usage. To ensure a fair algorithmic comparison with models like TConstFormer, the baseline model in this paper does not employ such additional engineering optimizations.

\paragraph{The Linear Advantage of TLinFormer and TConstFormer:}
In contrast, both of our proposed new architectures demonstrate exceptional scalability. TLinFormer (Figure~\ref{fig:inference_analysis}(b)), with its unique dual-mode performance characteristics, successfully constrains both the \textbf{upper bound (cache miss)} and \textbf{lower bound (cache hit)} of inference latency to a low-slope linear growth, significantly outperforming the baseline.

TConstFormer (Figure~\ref{fig:inference_analysis}(c)) achieves an even more fundamental breakthrough. Its performance upper bound also exhibits efficient linear growth, but its performance \textbf{lower bound} (the trough during a cache hit) is completely horizontal, no longer changing with sequence length $N$, demonstrating the ideal \textbf{constant-time $\mathcal{O}(1)$ characteristic}. This observation is perfectly consistent with our theoretical analysis (see Section~\ref{sec:complexity}) and fundamentally establishes TConstFormer's architectural advantage in long-sequence inference scenarios.

\subsubsection{KV Cache Mechanism Efficiency Comparison (Figures d, e, f):}

To quantify the actual benefits of caching, we compared the inference speedup ratio of each model during cache hits versus misses.

\noindent\textbf{Cache Failure of the Baseline Model.} As shown in Figure~\ref{fig:inference_analysis}(d), the speedup ratio of the standard model peaks at only 1.26x and rapidly decays to 1.0 as the sequence grows, confirming that in long-sequence scenarios, its naive KV cache mechanism \textbf{completely fails} due to the memory bandwidth bottleneck.

\noindent\textbf{Efficient Caching of TLinFormer and TConstFormer.} As shown in Figures~\ref{fig:inference_analysis}(e) and (f), the cache speedup ratios of both TLinFormer and TConstFormer show a strong positive correlation with sequence length, indicating that the longer the context, the more significant the performance gain from their caching mechanisms. TLinFormer achieves over a \textbf{10x} speedup on million-token sequences, while TConstFormer, with its constant-time cache hit cost, achieves an astonishing peak speedup of over \textbf{40x}, demonstrating a massive advantage.

\subsubsection{Memory Usage and Supported Sequence Length (Figure g)}

Figure~\ref{fig:inference_analysis}(g) provides a direct comparison of the cache memory usage of each model. The memory usage of both the baseline and TLinFormer grows linearly with sequence length $N$, although the latter has a gentler slope. TConstFormer once again demonstrates its fundamental architectural advantage, achieving \textbf{constant-level $\mathcal{O}(1)$ memory usage}, completely breaking free from the constraints of sequence length on memory.

\subsubsection{Overall Inference Speedup Ratio (Figures h, i)}

\begin{figure}[H]
    \centering

    \begin{subfigure}[b]{0.3\textwidth}
        \centering
        \includegraphics[width=\linewidth]{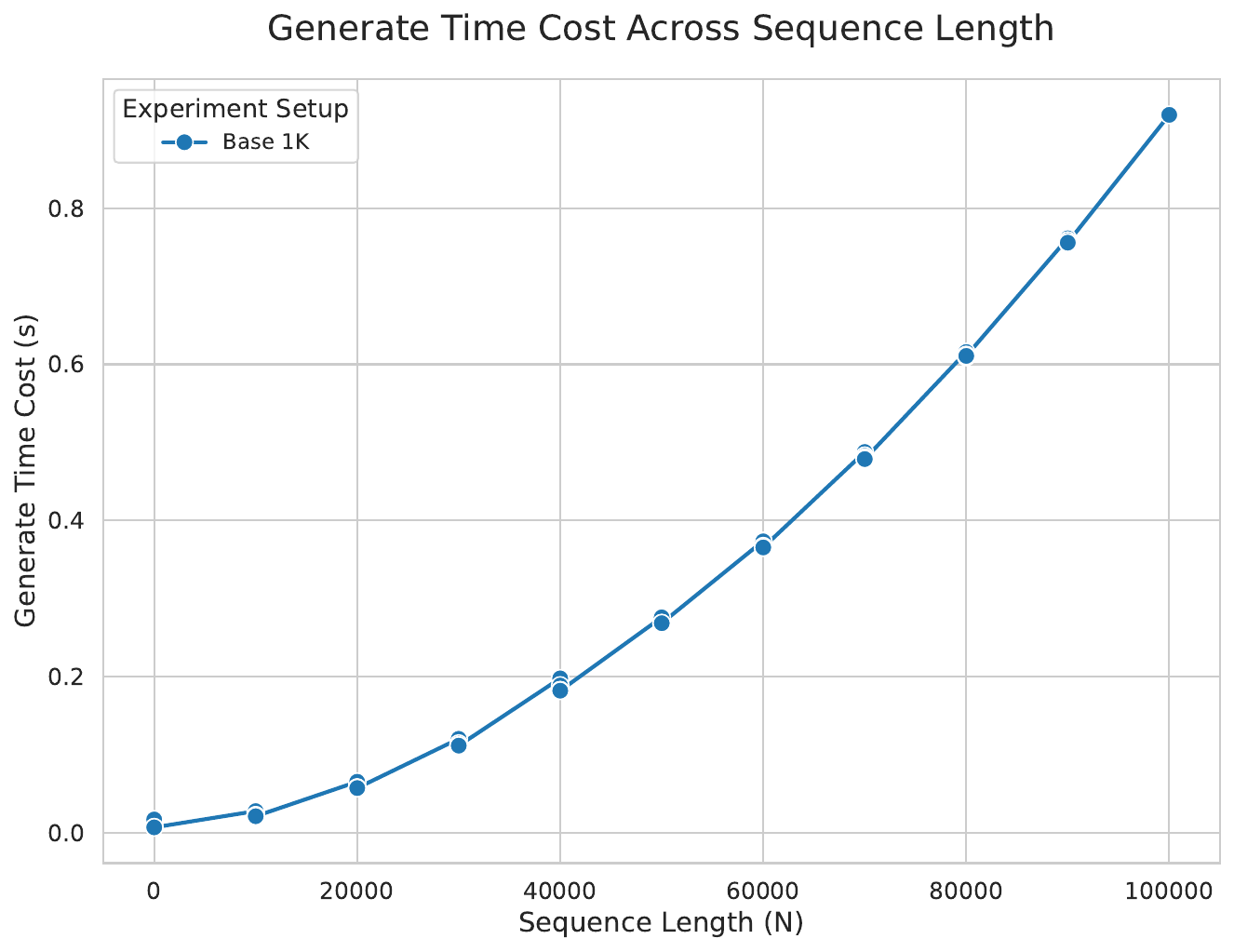}
        \caption{Baseline model inference time}
        \label{subfig:baseline_latency_inf}
    \end{subfigure}%
    \hfill
    \begin{subfigure}[b]{0.3\textwidth}
        \centering
        \includegraphics[width=\linewidth]{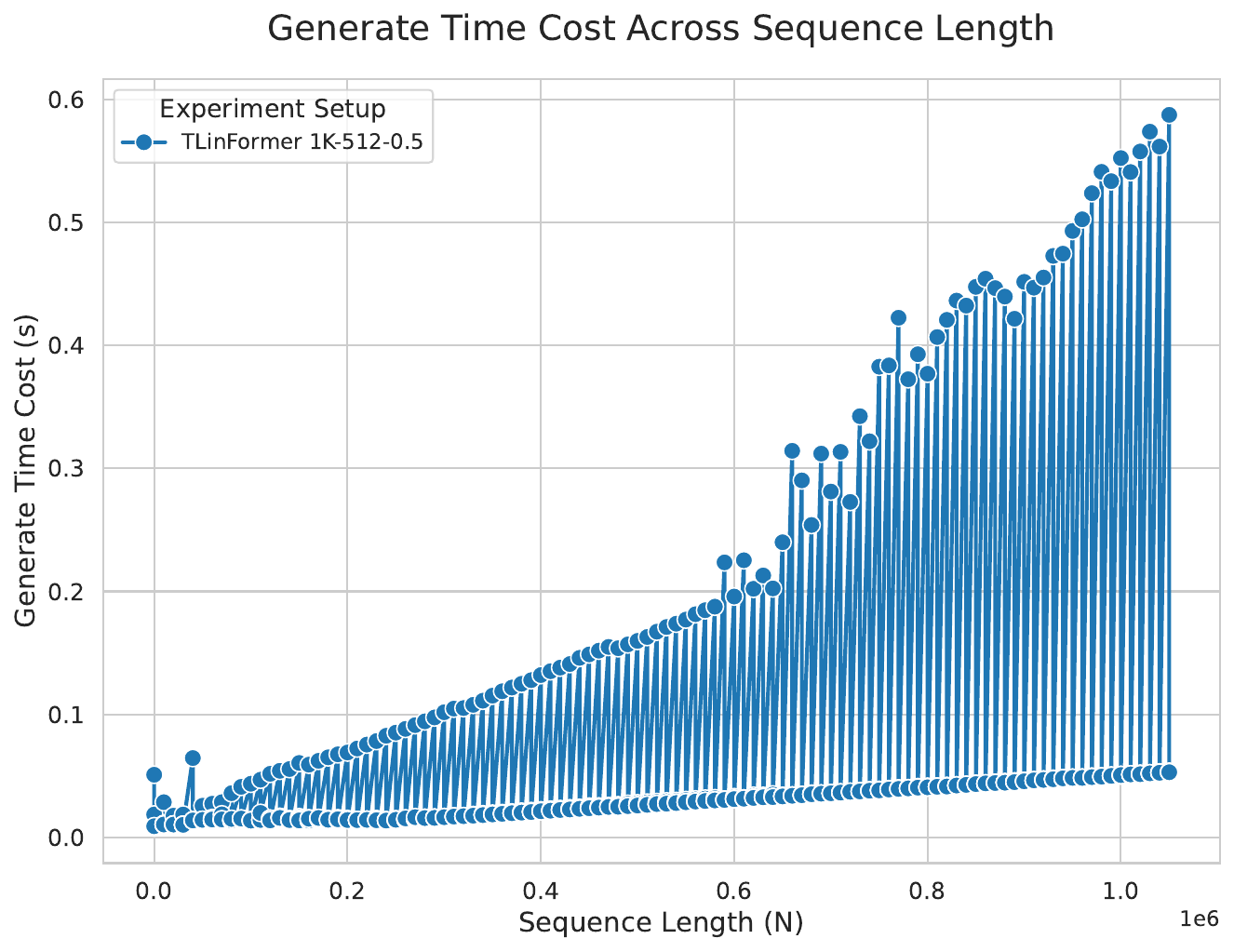}
        \caption{TLinFormer inference time}
        \label{subfig:tlinformer_latency_inf}
    \end{subfigure}
    \hfill
    \begin{subfigure}[b]{0.3\textwidth}
        \centering
        \includegraphics[width=\linewidth]{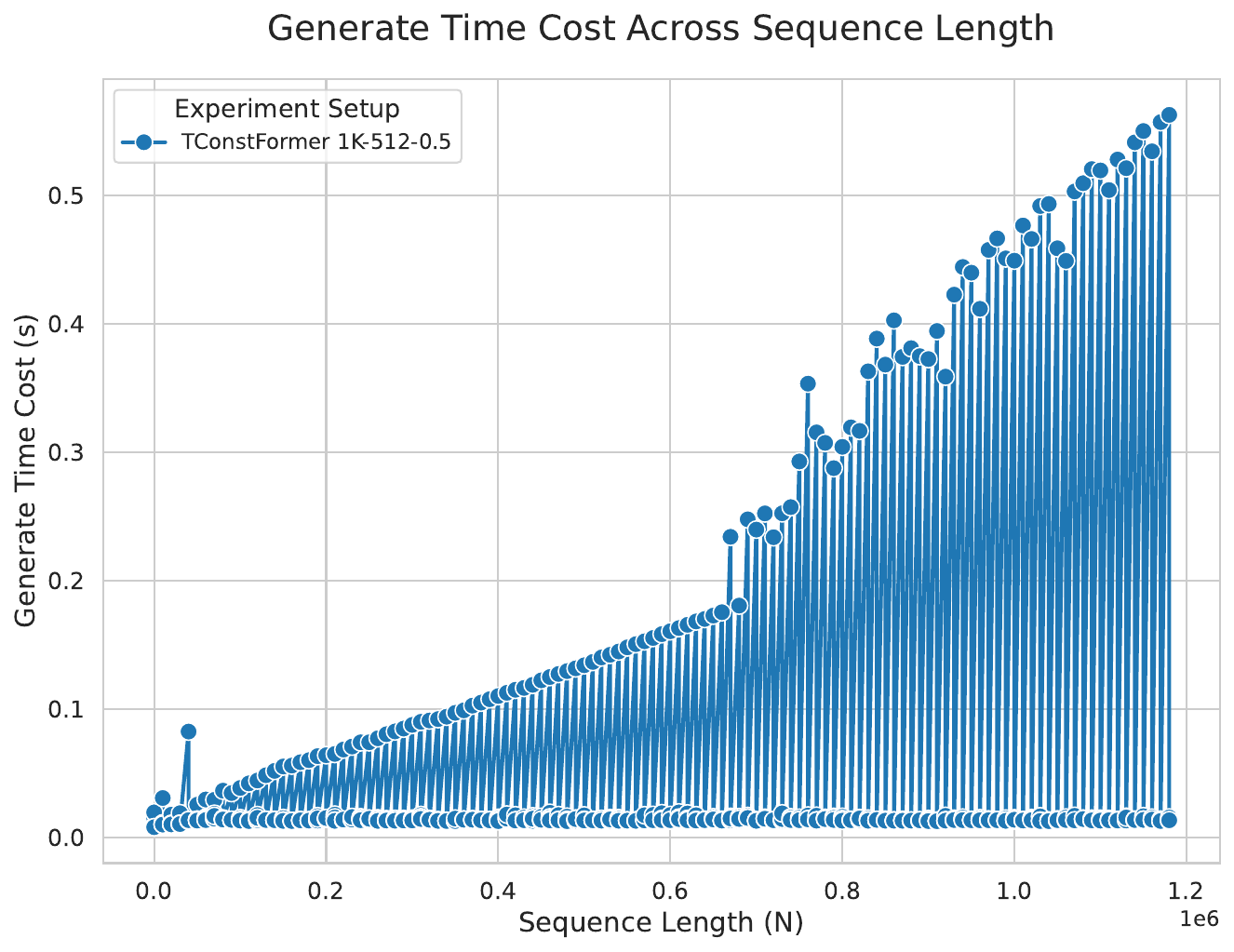}
        \caption{TConstFormer inference time}
        \label{subfig:tconstformer_latency_inf}
    \end{subfigure}

    \vspace{1.5em}

    \begin{subfigure}[b]{0.3\textwidth}
        \centering
        \includegraphics[width=\linewidth]{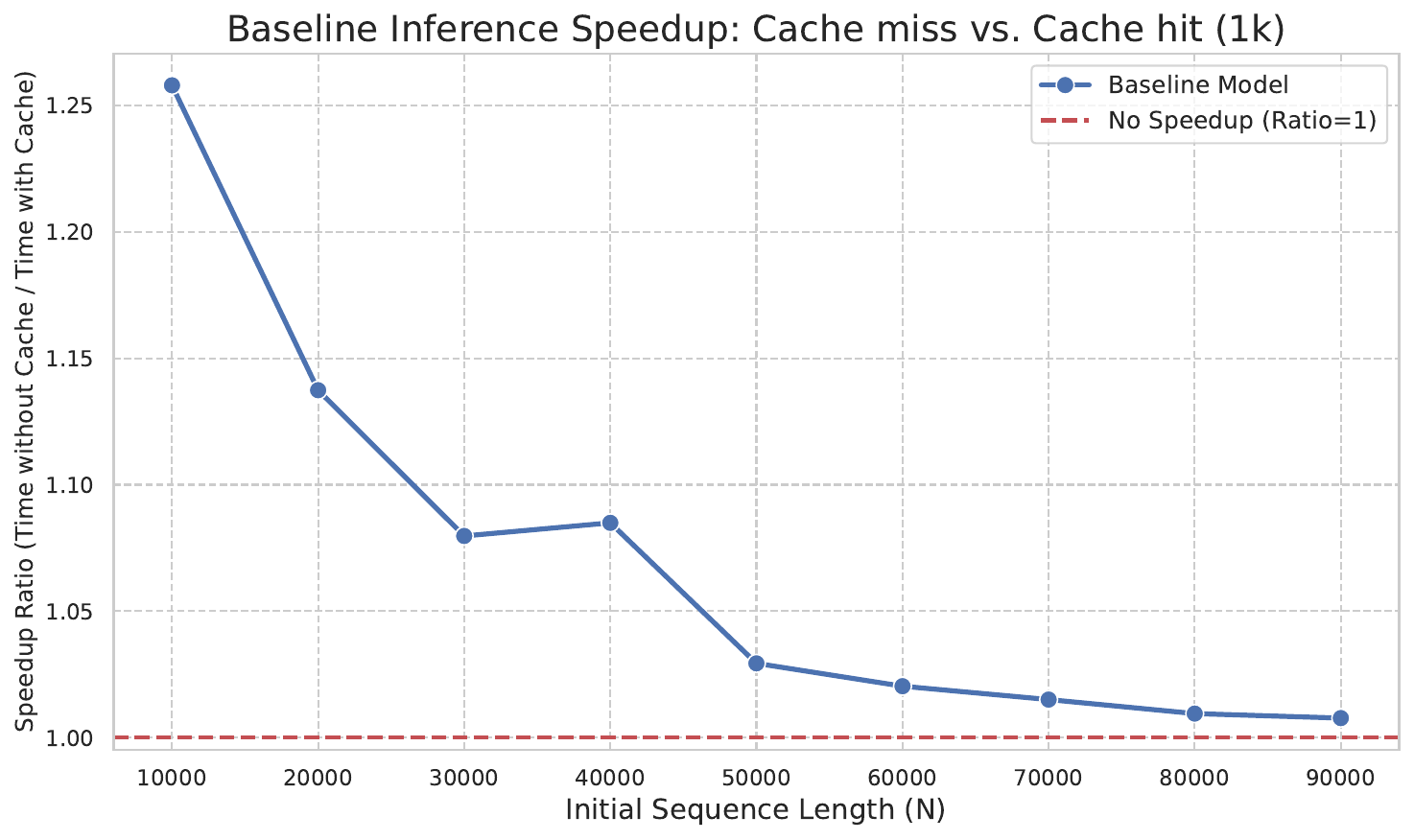}
        \caption{Baseline cache hit speedup}
        \label{subfig:baseline_speedup}
    \end{subfigure}%
    \hfill
    \begin{subfigure}[b]{0.3\textwidth}
        \centering
        \includegraphics[width=\linewidth]{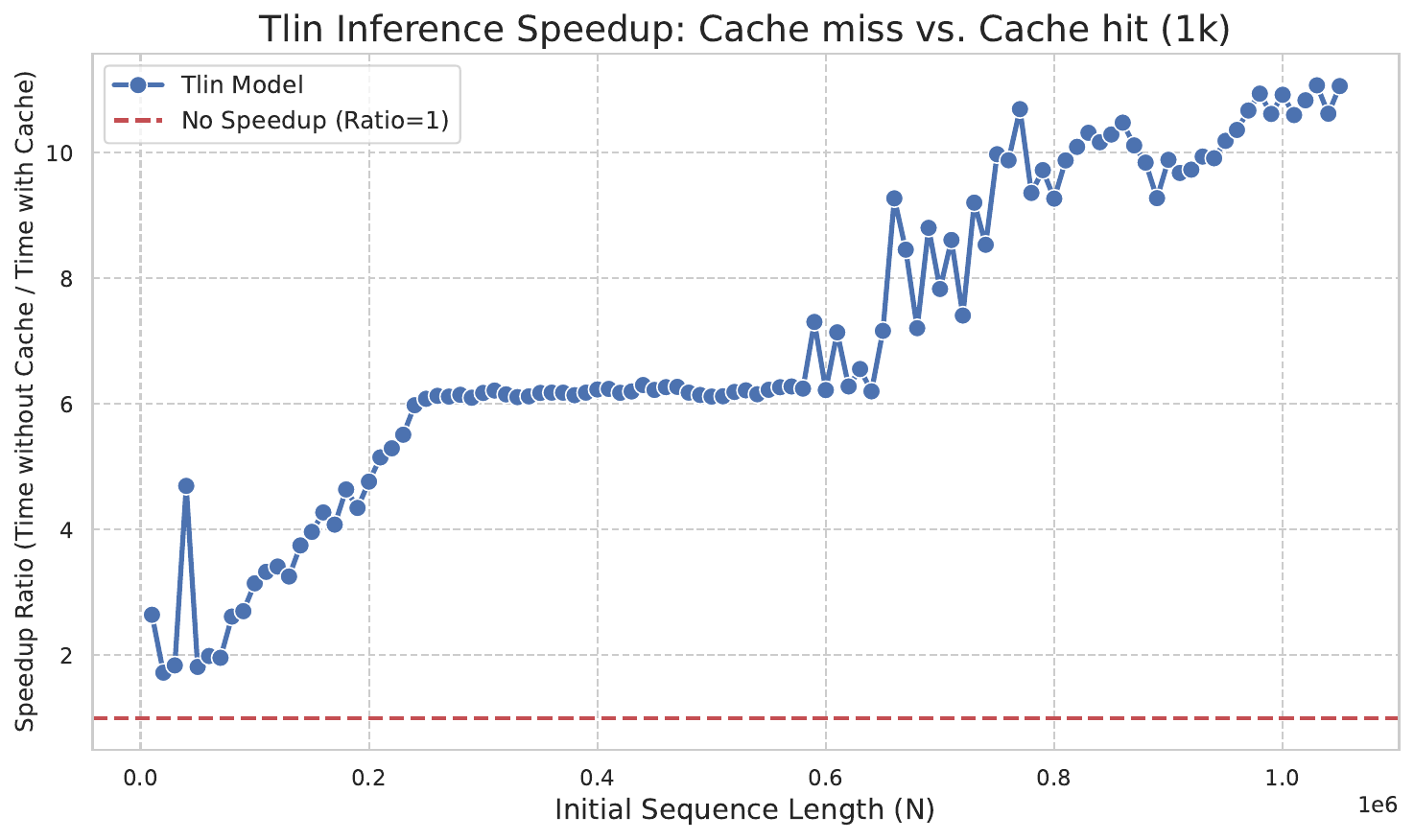}
        \caption{TLinFormer cache hit speedup}
        \label{subfig:tlin_speedup}
    \end{subfigure}
    \hfill
    \begin{subfigure}[b]{0.3\textwidth}
        \centering
        \includegraphics[width=\linewidth]{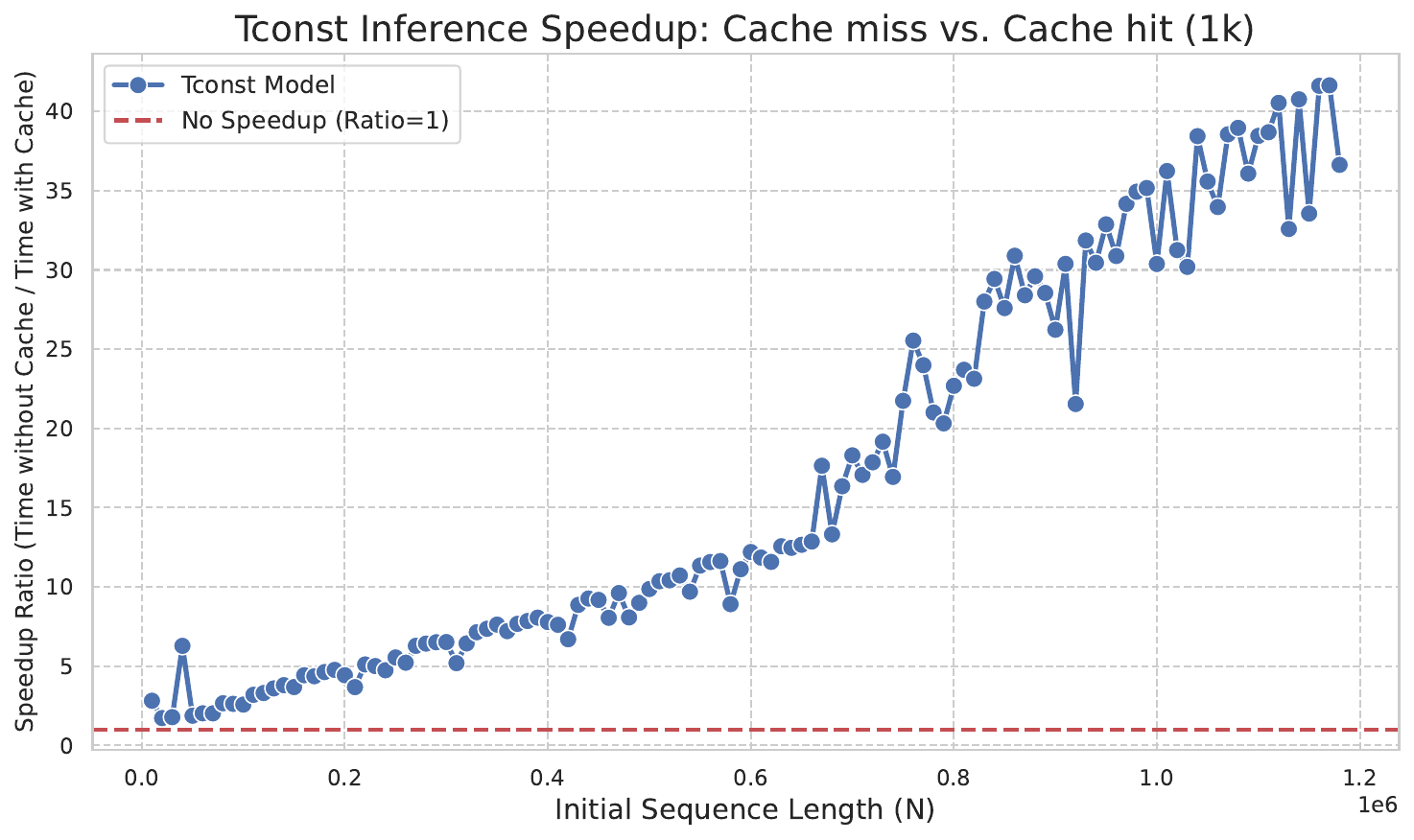}
        \caption{TConstFormer cache hit speedup}
        \label{subfig:tconst_speedup}
    \end{subfigure}

    \vspace{1.5em}

    \begin{subfigure}[b]{0.3\textwidth}
        \centering
        \includegraphics[width=\linewidth]{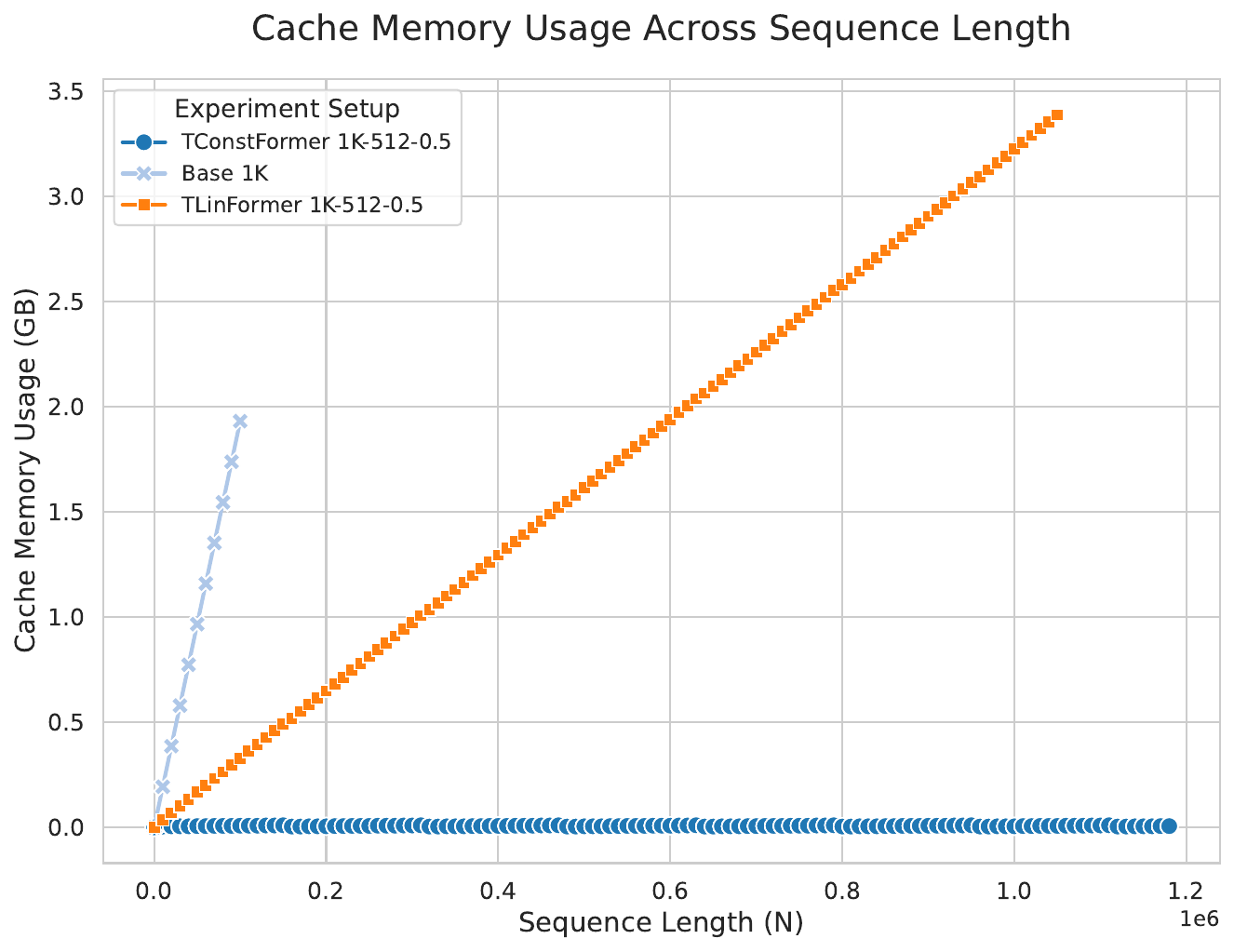}
        \caption{Cache memory usage of models}
        \label{subfig:cache_memory_usage}
    \end{subfigure}%
    \hfill
    \begin{subfigure}[b]{0.3\textwidth}
        \centering
        \includegraphics[width=\linewidth]{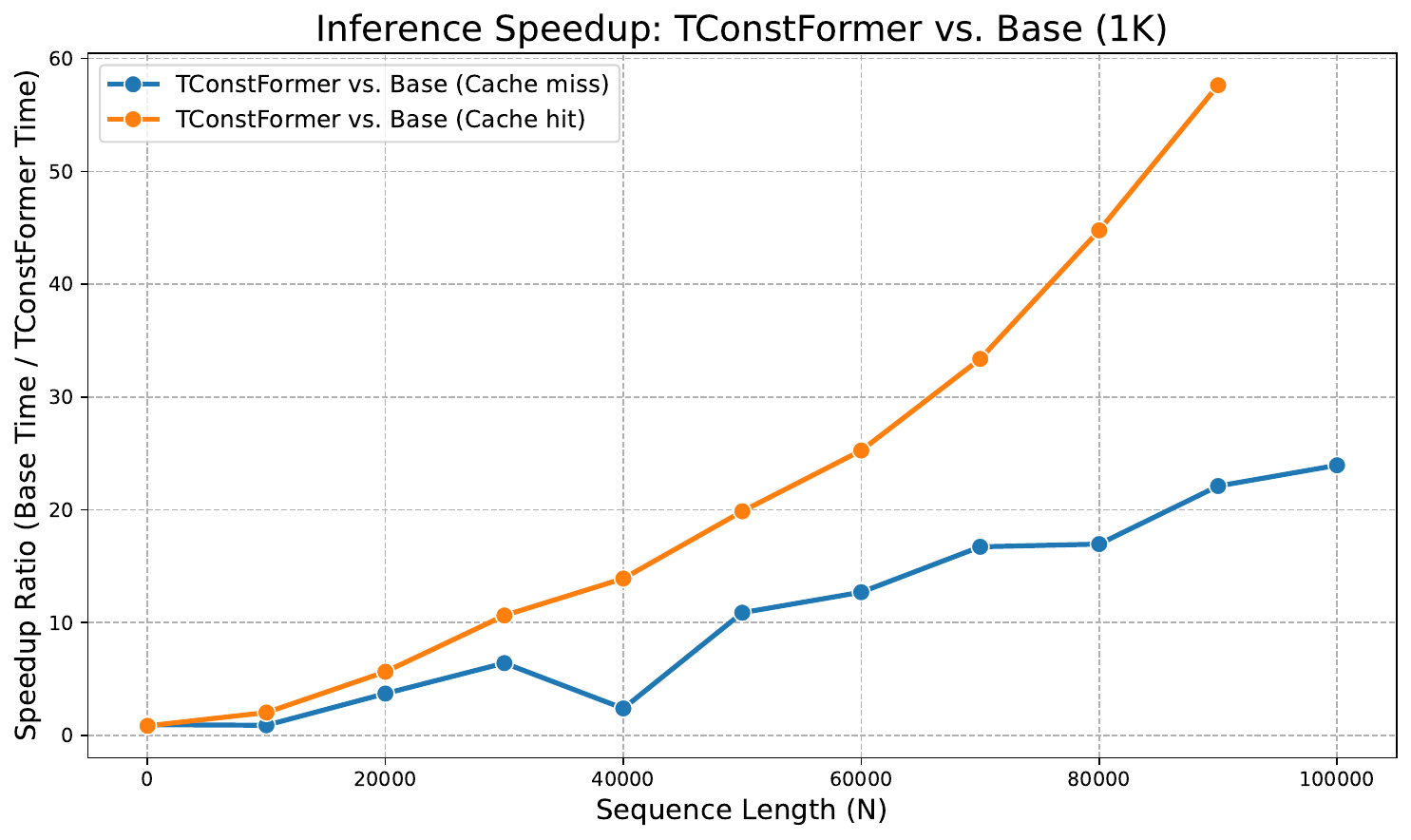}
        \caption{TConstFormer vs. Baseline time ratio}
        \label{subfig:tconst_base_speedup}
    \end{subfigure}
    \hfill
    \begin{subfigure}[b]{0.3\textwidth}
        \centering
        \includegraphics[width=\linewidth]{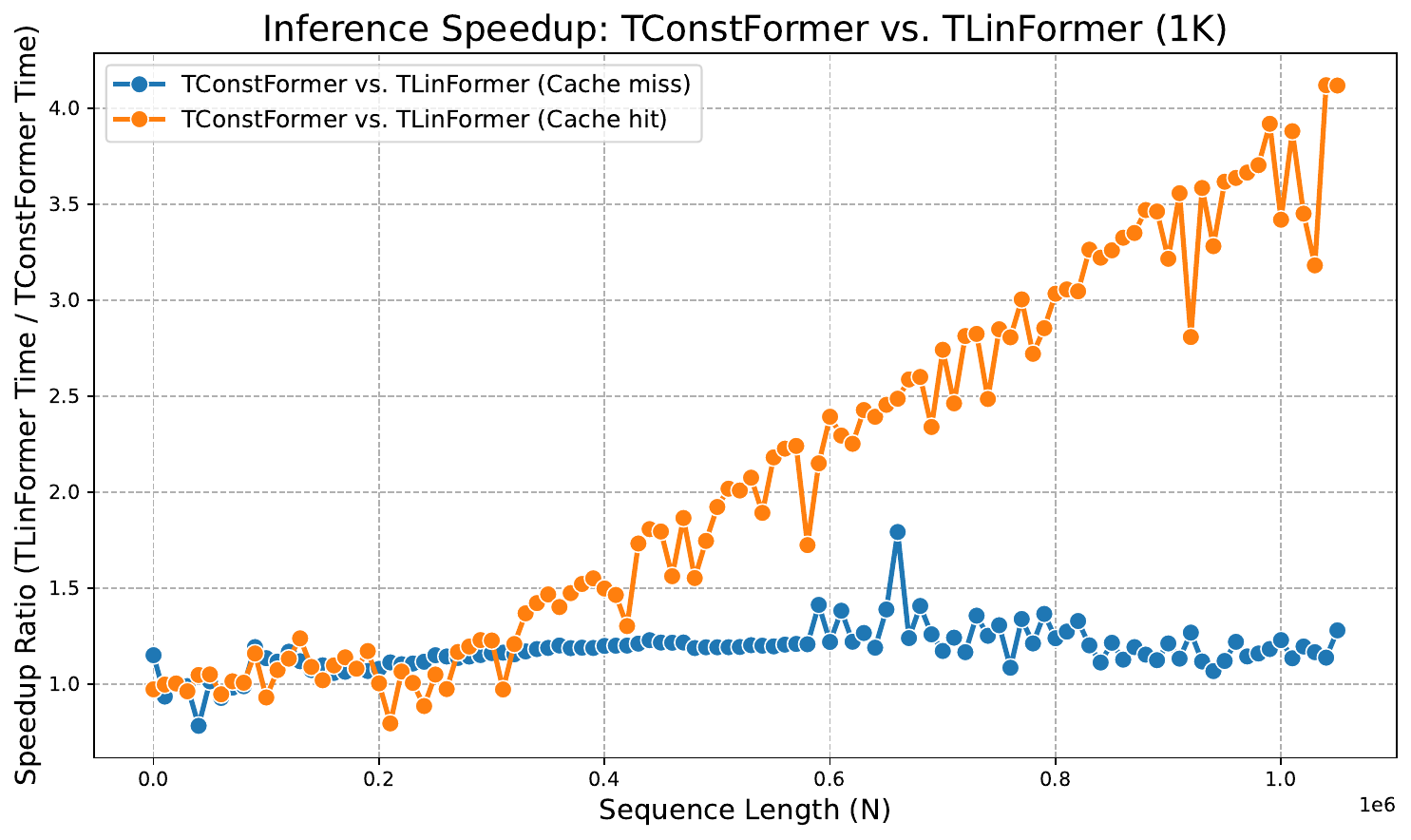}
        \caption{TConstFormer vs. TLinFormer time ratio}
        \label{subfig:tlin_tconst_speedup}
    \end{subfigure}

    \caption{
        \textbf{Inference performance and cache efficiency comparison.}
        (\textbf{a}) Baseline model's latency grows super-linearly with sequence length.
        (\textbf{b}, \textbf{c}) Both TLinFormer and TConstFormer demonstrate excellent scalability. Their dual-mode performance (peaks for cache miss upper bound, troughs for cache hit lower bound) clearly validates the cache mechanism's effectiveness, with TConstFormer's lower bound being constant.
        (\textbf{d}) Due to a memory bandwidth bottleneck (caused by \texttt{torch.cat}), the baseline's cache speedup ratio rapidly decays towards 1 (ineffective) as sequence length increases.
        (\textbf{e}, \textbf{f}) In contrast, the cache speedup ratios of TLinFormer and TConstFormer grow steadily with sequence length, showing significant and sustained acceleration.
        (\textbf{g}) In terms of cache memory usage, both new architectures are far below the baseline, supporting longer sequences, with TConstFormer achieving O(1) cache consumption.
        (\textbf{h}, \textbf{i}) Overall inference time comparison shows that TConstFormer achieves orders-of-magnitude speedup over the baseline and outperforms TLinFormer.
    }

    \label{fig:inference_analysis}
\end{figure}

Figures~\ref{fig:inference_analysis}(h) and (i) quantify the enormous advantage of TConstFormer from an end-to-end perspective.
\begin{itemize}
    \item \textbf{Compared to the Baseline Model (Figure h):} During a cache hit, TConstFormer achieves a speedup ratio that grows linearly with sequence length, reaching up to \textbf{tens of times} faster, representing an order-of-magnitude performance leap.
    \item \textbf{Compared to TLinFormer (Figure i):} Even when compared to the optimized TLinFormer, TConstFormer still achieves a significant and continuously growing performance advantage during a cache hit.
\end{itemize}

\subsection{Conclusion}

In summary, the experimental results collectively validate that TConstFormer is not just an effective improvement, but a solution with an overwhelming advantage in both time and space efficiency for long-sequence inference tasks.

\section{From Efficient Compression to Constant State: TConstFormer and Emergent Intelligence}
\label{sec:compression_to_emergence}

First, let's consider the limitations of the standard auto-regressive architecture. When we view a decoder-only architecture from a fully-connected perspective, it can be understood as the model internally creating an observation window that expands equivalently as the sequence length increases. This processing logic is unreasonable because for an infinitely long sequence, the model would need to generate an infinitely long internal observation window. External hardware constraints mean it must have an upper limit on the sequence length it can handle. This leads to an inevitable conclusion: \textbf{for a truly efficient intelligent agent, its understanding of the world (i.e., its internal state) must be completely decoupled from the length of its history in terms of computational and storage resources}. Therefore, the observation window for this input sequence must be bounded.

Second, since compression is necessary, what is the constraint between the compressed length and the original length? This is what we will explore next. Foundational principles from information theory and compressed sensing, notably the empirical guideline \( n > C \log N \), establish that high-dimensional signals often reside on a low-dimensional manifold~\cite{candes_robust_2004, abo-zahhad_compressive_2015}. Here, \(n\) is the dimension of the compressed representation required to faithfully reconstruct a signal of original dimension \(N\). This theoretical underpinning suggests that for processing long sequences, a remarkably small context window can be sufficient. For instance, to capture the essential information of a sequence with \(N = 10^7\) tokens, a compressed representation of dimension \( n \approx 134 \) (for \( C \approx 8.33 \)) could theoretically suffice.

Our TConstFormer architecture is precisely a solution to the aforementioned constraint. It not only inherits the "forced compression" philosophy of TLinFormer but elevates it to a new level. By achieving \textbf{constant-time $\mathcal{O}(1)$ cache updates and memory footprint}, TConstFormer implements architecturally what we call a \textbf{"Constant-State Representation"} mechanism.

This means that whether the historical sequence is one thousand, one million, or one billion tokens long, the "high-information-density state" that TConstFormer relies on to generate the next token is \textbf{completely invariant} in terms of storage cost and amortized computation. The model is thoroughly deprived of the ability to use "history length" as a shortcut. It is \textbf{forced} to learn a \textbf{scale-invariant knowledge distillation capability}—to indiscriminately refine historical information of any length into an internal state of fixed complexity that represents the core regularities of the world.

We speculate that this \textbf{Physical Constraint} on the complexity of the internal state may be a key prerequisite for the emergence of intelligence, especially general intelligence. Just as the human brain processes an infinite amount of information from the real world within a finite volume and energy budget, a truly general AI must also learn to model and predict an infinite stream of information with constant resource expenditure.

The constant-level efficiency of TConstFormer is more like a "golden hoop" imposed on the intelligent agent, one that aligns with the laws of the physical world. It is this very constraint that compels the model to move beyond its dependency on sequence length and learn deeper, more fundamental abstract principles.

Therefore, we believe that TConstFormer is not only a major step forward in the efficiency of long-sequence modeling but also a meaningful attempt to explore the \textbf{computational essence of intelligence}. Its "Constant-State" design philosophy gives us a clearer and more exciting glimpse of the path toward AGI.

\section{Limitations and Discussion}
\label{sec:limitations}

Although TConstFormer demonstrates significant efficiency advantages both theoretically and experimentally, we must acknowledge that this study has several limitations, which also point toward directions for future work.

\paragraph{Model Scale and Task Complexity.}
The experimental validation in this study was primarily conducted on a small-scale model with approximately 41M parameters, largely due to the computational resource constraints of individual research. Consequently, the performance of TConstFormer on large language models at the scale of billions of parameters, as well as its ability to replicate the emergent capabilities in complex tasks (such as long-context instruction following), remains an open question. Scaling the TConstFormer architecture to larger models is a critical next step to validate its effectiveness in real-world, complex applications.

\paragraph{On the Capability for Precise Information Retrieval.}
The 'Constant-State' mechanism of TConstFormer essentially compresses and distills historical information. While this mechanism is advantageous for learning macroscopic semantics and structural patterns in text, its performance on tasks requiring \textbf{verbatim recall} (such as the 'Needle in a Haystack' test) is an area where we have not yet obtained definitive data, due to the limited scale of our model. This also represents a highly valuable direction for future research.


\section{Conclusion and Future Work}
\label{sec:conclusion}

In this paper, we departed from the mainstream paradigm of attention approximation and returned to the first principles of connectionism, proposing a constant-attention architecture—TConstFormer—from the perspective of information flow topology that achieves a fundamental breakthrough in efficiency. By identifying and reconstructing the performance-bottleneck pathways in TLinFormer, TConstFormer inherits its \textbf{computational precision} and \textbf{full-context accessibility} while, for the first time, reducing both the computational complexity and KV cache overhead of autoregressive inference to a \textbf{constant level ($\mathcal{O}(1)$)}. Experiments demonstrate that TConstFormer provides an extremely efficient, robust, and scalable solution for long-sequence modeling, thereby significantly lowering the hardware barrier for ultra-long sequence applications.

As discussed in Section~\ref{sec:limitations}, while this study has limitations regarding model scale, the "Constant-State" architecture of TConstFormer opens up several exciting directions for future AI model design:
\begin{itemize}
    \item \textbf{Integration with Cutting-Edge Technologies:} TConstFormer's constant-level efficiency is orthogonal to and highly compatible with parameter-efficient optimization techniques such as Mixture-of-Experts (MoE). Combining them holds the promise of building next-generation foundation models that reach new heights in both parameter scale and performance, all within a limited computational budget.
    
    \item \textbf{Towards Infinite Context and Streaming Processing:} The $\mathcal{O}(1)$ inference cost means TConstFormer is naturally suited for \textbf{streaming processing of infinitely long sequences}, such as handling real-time video streams, unending dialogues, or continuous sensor data. Exploring its application in such open-ended tasks is a highly promising direction.
    
    \item \textbf{Exploring Higher-Dimensional Tensorial Attention:} The core idea of this paper leads to a more profound question: can we generalize this efficiency optimization, based on connection topology, from sequences (L) to higher-dimensional tensor data (e.g., video data \texttt{[T, H, W, C]})? Developing computationally feasible High-dimensional Tensorial Attention could be a key step towards more general and powerful AI models.
\end{itemize}

Finally, in Section~\ref{sec:compression_to_emergence}, we discussed the necessity of compression and pointed out that TConstFormer is precisely a solution that embodies this philosophy.

\section*{Code Availability}
The source code for this paper is available at \url{https://github.com/simonFelix-Ai/TConstFormer}. The code is dual-licensed for academic and commercial use.

\bibliographystyle{plain}
\bibliography{references} 

\appendix 
\section{Detailed Derivation of Computational Complexity}
\label{app:complexity_derivation}

This appendix provides a detailed derivation of Equations~\eqref{eq:total_cost_simple_cache_miss} and~\eqref{eq:total_cost_simple_cache_hit} from the main text. We analyze the upper bound of the computational cost by considering a full computation cycle where both the context and generation windows are updated. The analysis is divided into costs associated with the left (context) and right (generation) windows.

\subsection{Cache Miss}

\begin{enumerate}
    \item \textbf{Computational Cost of the Left Window (Historical Context)}:
    \begin{itemize}
        \item \textbf{First Layer Cross-Attention}: The query sequence from the context window attends to the full history. Cost is $D \cdot (N - \Wog) \cdot \Woh$.
        \item \textbf{Intermediate Self-Attention Layers ($H$ layers)}: Self-attention is performed within the context window of size $\Woh$. Cost is $H \cdot D \cdot \Woh^2$.
        \item \textbf{Final Layer Cross-Attention (Dimension Restoration)}: The full history attends to the processed context window to restore the original sequence length. Cost is $D \cdot (N - \Wog) \cdot \Woh$.
        \item \textit{\textbf{Total Cost of the Left Window ($C_{\text{left}}$):}}
        \[
        C_{\text{left}} = 2 D (N - \Wog) \Woh + H D \Woh^2
        \]
    \end{itemize}
\item \textbf{Computational Cost of the Right Window (Generation Area)}:
\begin{itemize}
    \item \textbf{Cross-Attention with Intermediate Context Layers ($H+1$ layers, including final output layer)}: The generation window attends to the processed context window. Cost is $(H + 1) \cdot D \cdot \Wog \cdot \Woh$.
    \item \textbf{Causal Self-Attention (All $H+2$ layers, including final output layer)}: Causal self-attention is performed within the generation window. Cost is $(H + 2) \cdot D \cdot \Wog^2$.
    \item \textit{\textbf{Total Cost of the Right Window ($C_{\text{right}}$):}}
    \[
    C_{\text{right}} = (H + 1) D \Wog \Woh + (H+2)D \Wog^2
    \]
\end{itemize}

\item \textbf{Derivation of Total Computational Cost ($T$):}
The total cost is the sum of the costs of the two windows, $T = C_{\text{left}} + C_{\text{right}}$.
\begin{align*}
    T &= \left[ 2 D (N - \Wog) \Woh + H D \Woh^2 \right] \\
      &\quad + \left[ (H + 1) D \Wog \Woh + (H+2)D \Wog^2 \right] \\
    \intertext{Step 1: Expand all terms}
    &= \left( 2DN\Woh - 2D\Wog\Woh + HD\Woh^2 \right) \\
      &\quad + \left( HD\Wog\Woh + D\Wog\Woh + HD\Wog^2 + 2D\Wog^2 \right) \\
    \intertext{Step 2: Combine like terms}
    &= 2DN\Woh - 2D\Wog\Woh + D\Wog\Woh + HD\Woh^2 + HD\Wog\Woh + HD\Wog^2 + 2D\Wog^2 \\
    &= 2DN\Woh - D\Wog\Woh + HD\Woh^2 + HD\Wog\Woh + HD\Wog^2 + 2D\Wog^2 \\
    \intertext{Step 3: Factor out common factor $D$ and reorganize by variables $N$ and $H$}
    &= D \left[ 2N\Woh - \Wog\Woh + H\Woh^2 + H\Wog\Woh + H\Wog^2 + 2\Wog^2 \right] \\
    \intertext{Step 4: Final organized form (grouping terms related to $N$ and $H$)}
    &= D \left[ N(2\Woh) + H(\Woh^2 + \Wog^2 + \Wog\Woh) + 2\Wog^2 - \Wog\Woh \right]
\end{align*}
    Derivation complete.
\end{enumerate}

\subsection{Cache Hit}

\begin{enumerate}
    \item \textbf{Computational Cost of the Left Window (Historical Context)}:
    \begin{itemize}
        \[
        C_{\text{left}} = 0
        \]
    \end{itemize}
\item \textbf{Computational Cost of the Right Window (Generation Area)}:
\begin{itemize}
    \item \textbf{Cross-Attention with Intermediate Context Layers ($H+1$ layers, including final output layer)}: Only the last token of the generation window participates in the computation. Cost is $(H + 1) \cdot D \cdot \Woh$.
    \item \textbf{Causal Self-Attention (All $H+2$ layers, including final output layer)}: Causal self-attention is performed within the generation window. The upper bound on cost is $(H + 2) \cdot D \cdot \Wog^2$.
    \item \textit{\textbf{Total Cost of the Right Window ($C_{\text{right}}$):}}
    \[
    C_{\text{right}} = (H + 1) D \Woh + (H+2)D \Wog^2
    \]
\end{itemize}

\item \textbf{Derivation of Total Computational Cost ($T$):}
The total cost is the sum of the costs of the two windows, $T = C_{\text{left}} + C_{\text{right}}$.
\begin{align*}
    T &= (H + 1) D \Woh + (H+2)D \Wog^2
\end{align*}
    Derivation complete.
\end{enumerate}

\end{document}